\documentclass{article} 
\usepackage{preprint,times}

\usepackage{amsmath,amsfonts,bm}

\def\eqref#1{equation~\ref{#1}}

\def\1{\bm{1}}

\DeclareMathAlphabet{\mathsfit}{\encodingdefault}{\sfdefault}{m}{sl}
\SetMathAlphabet{\mathsfit}{bold}{\encodingdefault}{\sfdefault}{bx}{n}

\usepackage{graphicx}
\usepackage{amsmath}
\usepackage{amssymb}
\usepackage{booktabs}
\usepackage{amsthm}
\usepackage{svg}
\usepackage{multirow}
\usepackage{multicol}
\usepackage{bm}
\usepackage{bbm}
\usepackage{xcolor}

\definecolor{samgreen}{RGB}{24,202,56}

\newcommand{\cov}[1]{\mathbf{\Sigma}\left[\mathbf{#1}\right]}
\newcommand{\var}[1]{\text{Var}\left[#1\right]}
\newcommand{\B}[1]{\textbf{#1}}
\newcommand{\ul}[1]{\underline{#1}}
\newtheorem{theorem}{Theorem}[section]

\DeclareMathOperator{\diag}{\textup{diag}}
\DeclareMathOperator{\diagembed}{\textup{diag-embed}}
\DeclareMathOperator{\trace}{\textup{Tr}}

\usepackage{hyperref}
\usepackage{url}

\title{PHI-S: Distribution Balancing for Label-Free Multi-Teacher Distillation}

\author{Mike Ranzinger, Jon Barker, Greg Heinrich, \\ \B{Pavlo Molchanov, Jan Kautz, Bryan Catanzaro, Andrew Tao} \\
NVIDIA\\
\texttt{mranzinger@nvidia.com} 
}

\arxivfinalcopy 
\begin{document}

\maketitle

\begin{figure*}[!ht]
    \centering\resizebox{1.0\linewidth}{!}{
        \includegraphics{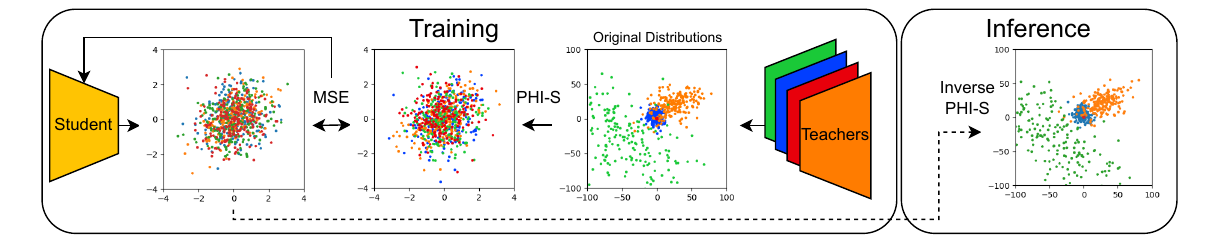}
    }
    \caption{
    Illustration of the modified agglomerative model training procedure. Instead of the student model learning to match the original teacher distributions, it learns to match the normalized distributions (our proposed PHI-S is shown). We show the real distributions for \textcolor{red}{DFN CLIP}, \textcolor{blue}{DINOv2}, \textcolor{orange}{SigLIP}, and \textcolor{samgreen}{SAM} by projecting them down to 2D using PCA. In the original space, the variance of DFN CLIP is so small that it appears as a single point. During inference, we can estimate the original teacher distributions using the inverse normalization process on the student predictions.
    }
    \label{fig:ship_teaser}
\end{figure*}

\begin{abstract} 
Various visual foundation models have distinct strengths and weaknesses, both of which can be improved through heterogeneous multi-teacher knowledge distillation without labels, termed ``agglomerative models.'' We build upon this body of work by studying the effect of the teachers' activation statistics, particularly the impact of the loss function on the resulting student model quality. We explore a standard toolkit of statistical normalization techniques to better align the different distributions and assess their effects. Further, we examine the impact on downstream teacher-matching metrics, which motivates the use of Hadamard matrices. With these matrices, we demonstrate useful properties, showing how they can be used for isotropic standardization, where each dimension of a multivariate distribution is standardized using the same scale. We call this technique ``PHI Standardization'' (PHI-S) and empirically demonstrate that it produces the best student model across the suite of methods studied.
\end{abstract}

\section{Introduction}
\label{sec:intro}

A body of work recently emerged on the topic of agglomerative models \cite{ranzinger2023amradio}, which is fusing multiple heterogeneous visual foundation models \cite{awais2023foundational} into a single model via multi-teacher knowledge distillation \cite{hinton2015distilling,zuchniak2023multiteacher} without labels. Starting with AM-RADIO~\cite{ranzinger2023amradio}, and followed by Theia~\cite{shang2024theia}, and UNIC~\cite{sariyildiz2024unicuniversalclassificationmodels}. Theia and UNIC apply feature standardization to the teacher output, and demonstrate how important it is.

While knowledge distillation has a large body of literature (e.g. \cite{bucilu2006model,ahn2019variational,heo2019overhaul,huang2017like,Romero2014FitNetsHF,sun2021dynamic,wei2022contrastive,zagoruyko2017paying}), agglomerative models - dealing with multiple teachers coming from different modeling domains (e.g. vision-language contrastive \cite{radford2021clip}, self-supervised learning \cite{oquab2023dinov2,zhou2022ibot,assran2023ijepa}, and segmentation \cite{kirillov2023sam}) without ground truth labels - was new territory. In AM-RADIO, the authors chose DFN CLIP~\cite{fang2023data}, DINOv2-g-reg~\cite{darcet2023vision}, and SAM~\cite{kirillov2023sam} as their teacher models. While the authors studied loss balancing between the different teachers to some degree, they landed on a simple balancing strategy which was to apply the same weight to each teacher, both for summary and feature losses, and to use a linear combination of Cosine and Smooth-L1~\cite{girshick2015fast} objectives for feature matching.

\begin{figure*}[!t]
    \centering
    \resizebox{\linewidth}{!}{
        \includegraphics{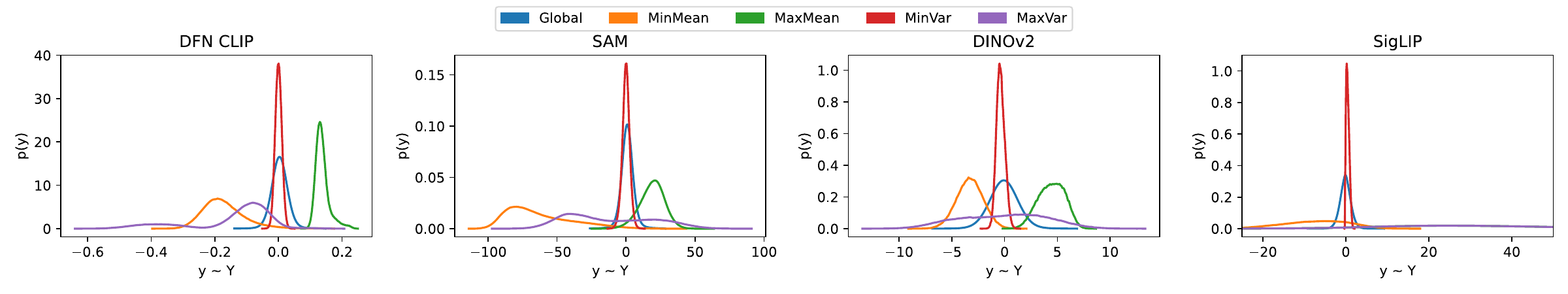}
    }
    \caption{Teacher activation histograms. We show the global histogram, as well as the histograms for the channels associated with the minimum mean, maximum mean, minimum variance, and maximum variance. While all being roughly normal, they have very different centers and scales. We provide specific values in table \ref{tab:teacher_activation_stats} in the appendix.}
    \label{fig:teacher_activation_stats}
\end{figure*}

In this work we study whether the choice of feature distillation loss function in AM-RADIO (equation 3) was an optimal choice. To motivate this, we start by analyzing the feature activation distributions for various teachers in figure \ref{fig:teacher_activation_stats}, and confirm that the distributions have very different variances. Notably, both Mean Squared Error (MSE) and Smooth-L1 are sensitive to variance scale, and thus, left uncontrolled for, each teacher will be implicitly weighted. For example, SAM's distribution has a standard deviation that is $191 \times$ larger than that of DFN CLIP. We also note that these distributions aren't a particularity of the training procedure by introducing SigLIP~\cite{zhai2023sigmoid} which has gained recent popularity due to its high scores on the OpenCLIP~\cite{ilharco2021openclip} leaderboard, as well as strong results within VLLMs \cite{fang2024vila2vilaaugmentedvila,li2024llavanextinterleavetacklingmultiimagevideo}. 

\noindent\paragraph{Main Contributions:}
\begin{itemize}
    \item We study the distributions of the teachers studied in \cite{ranzinger2023amradio} (plus SigLIP).
    \item We employ a statistical toolkit of standardization and whitening, and study their effects on downstream metrics.
    \item We study the effects of rotation matrices when applying whitening after identifying that the orientation of the normalized teacher distribution may affect the student model's errors.
    \item We study an application of Hadamard matrices on both whitening and standardization.
    \item In the case of standardization, we demonstrate that the Hadamard matrix may be used to produce a distribution that is standardized using a uniform scale across dimensions. We call this normalization method ``PHI Standardization'' (PHI-S) and demonstrate that it produces the best student models across our evaluation suite. 
\end{itemize}

\begin{figure*}[!h]
    \centering\resizebox{1.0\linewidth}{!}{
        \includegraphics{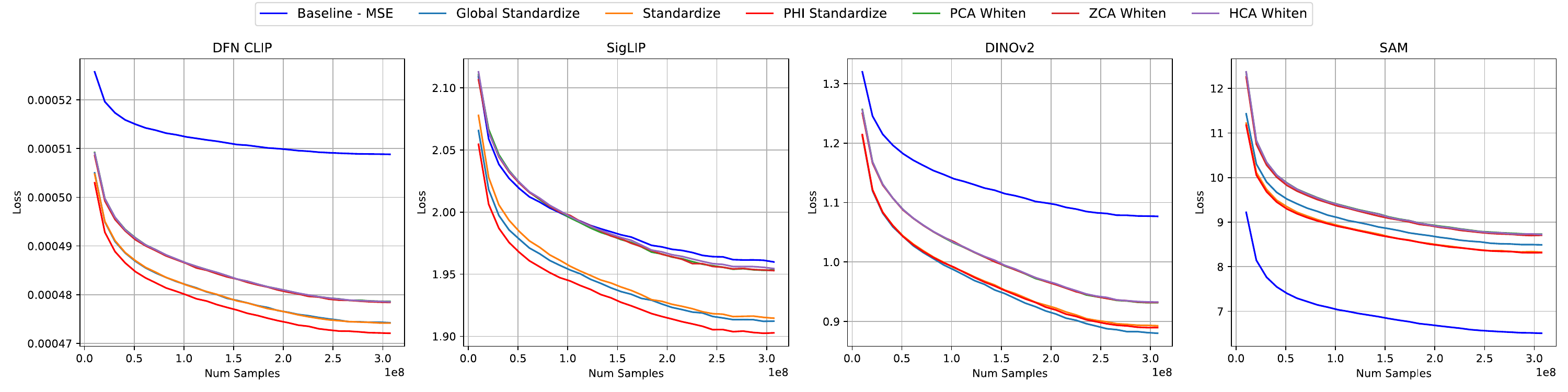}
    }
    \caption{
    The loss curves for each of the four teachers that the ViT-B/16 student is learning to match simultaneously in original teacher space (e.g. denormalized). We emphasize ``Baseline - MSE'' (Blue) and ``PHI Standardize'' (PHI-S, Red) as they generally set the upper and lower bounds.
    }
    \label{fig:loss_plots}
\end{figure*}

\begin{table*}[!h]
    \centering
    \resizebox{\linewidth}{!}{
    \begin{tabular}{r|cccccc|ccc}
                         \multirow{2}{*}{\B{Method}} & \B{Teacher}     & \B{Classif-}  & \B{Segment-}& \B{SAM}  & \B{LLaVA}& \B{Probe}& \multirow{2}{*}{\B{Avg}}  & \B{Avg}     & \B{Avg No}   \\
                          & \B{MSE}                                    & \B{ication}   & \B{ation}   & \B{COCO} & \B{1.5}  & \B{3D}   &                           & \B{No COCO} & \B{MSE/COCO} \\
                     \hline
                   & \multicolumn{9}{c}{\B{Baselines}}                                                                                                            \\
                     \hline
        MSE               & 6.25                                       & 10.00         & 10.00       & \B{1.00} & 9.75     & 10.00    & 7.83                      & 9.20        & 9.94        \\
        Cosine            & 10.00                                      & \B{1.00}      & \B{2.00}    & 8.00     & 6.63     & 7.25     & 5.81                      & 5.38        & 4.22        \\
        Hyb MSE           & 5.50                                       & 9.00          & 3.00        & \ul{2.00}& 7.63     & 7.50     & 5.77                      & 6.53        & 6.78        \\
        Hyb SmL1          & 7.00                                       & \ul{3.00}     & 4.50        & 7.00     & 4.38     & 6.00     & 5.31                      & 4.98        & 4.47        \\
                    \hline   
                    & \multicolumn{9}{c}{\B{Standardization}}                                                                                                       \\
                    \hline
        \ul{Global Stdze} & \ul{2.75}                                  & \ul{3.00}     & 4.50        & 5.00     & \ul{3.13}& 4.50     & \ul{3.81}                 & \ul{3.58}   & \ul{3.78}   \\
        Standardize       & \ul{2.75}                                  & 5.00          & 7.50        & 3.00     & \B{2.75} & 4.25     & 4.21                      & 4.45        & 4.88        \\
        \B{PHI-S}          & \B{2.00}                                   & \ul{3.00}     & \B{2.00}    & 4.00     & 3.50     & \B{3.00} & \B{2.92}                  & \B{2.70}    & \B{2.88}    \\
                    \hline   
                    & \multicolumn{9}{c}{\B{Whitening}}                                                                                                             \\
                    \hline
        PCA-W             & 6.50                                       & 7.50          & 7.50        & 6.00     & 6.50     & \ul{3.75}& 6.29                      & 6.35        & 6.31        \\
        ZCA               & 5.50                                       & 7.50          & 8.00        & 9.00     & 4.75     & 4.00     & 6.46                      & 5.95        & 6.06        \\
        HCA               & 6.75                                       & 6.00          & 6.00        & 10.00    & 6.00     & 4.75     & 6.58                      & 5.90        & 5.69              
    \end{tabular}
    }
    \caption{Average ordinal rank between methods (1 is best, 10 is worst) across the benchmark suite for ViT-B/16. We observe that the standardization techniques work the best, with PHI-S being the strongest normalization method studied. The raw benchmark scores are provided in appendix \ref{sec:vitb_raw_scores}.}
    \label{tab:vitb_method_ranks}
\end{table*}

\section{Method}

The goal of the agglomerative student model is to produce activations $\mathbf{x}^{(t)}$ that match the teacher activations $\mathbf{y}^{(t)} \in \mathbf{Y}^{(t)}$ as closely as possible for each teacher $t \in T$, and the loss is (usually) linearly aggregated using weights $\alpha^{(t)}$. Finding these $\alpha^{(t)}$ is difficult due to the size of the design space, so current methods typically default to $\alpha^{(t)} = 1$ and focus on conditioning $\mathbf{Y}^{(t)}$ to handle distributional differences. For simplicity, we drop the $\cdot^{(t)}$ superscript as the same type of normalization is applied for every teacher, and each teacher has independent normalization parameters. Throughout this paper, we refer to $\var{\mathbf{Z}}$ as the diagonal of the covariance matrix $\cov{\mathbf{Z}}$ for some distribution $\mathbf{Z}$.

\subsection{Baseline}

We start with the MSE (mean squared error) loss serving as the baseline for feature matching:

\begin{align}
    L_{\text{mse}}(\mathbf{x}, \mathbf{y}) = \frac{1}{N} \sum_{n=1}^{N}\left(\mathbf{x}_n - \mathbf{y}_n \right)^2
    \label{eq:mse}
\end{align}

 Because AM-RADIO~\cite{ranzinger2023amradio} doesn't use MSE as their loss function, but rather a hybrid cosine + Smooth-L1 loss, we also consider a few of these variants. For example, the vanilla cosine distance loss, which is identical to what AM-RADIO uses for the summary loss. While we expect this to do poorly on the task of exactly matching the teacher distribution (due to magnitude invariance), it's not clear how this will affect the downstream tasks, so we include it.

\begin{equation}
    L_{\text{cos}}(\mathbf{x}, \mathbf{y}) = \frac{1}{N} \sum_{n=1}^{N} \left( 1 - \frac{\mathbf{x}^\intercal \mathbf{y}}{\left\lVert\mathbf{x}\right\rVert \left\lVert\mathbf{y}\right\rVert}\right)
    \label{eq:cos}
\end{equation}

We also consider the exact loss function proposed in AM-RADIO which is a hybrid of cosine distance and smooth-L1:

\begin{equation}
    L_{\text{hyb-sml1}}(\mathbf{x}, \mathbf{y}) = \beta \cdot L_{\text{cos}}(\mathbf{x}, \mathbf{y}) + (1-\beta) \cdot \text{SmoothL1}(\mathbf{x}, \mathbf{y})
    \label{eq:hyb_sml1}
\end{equation}

For completeness, we ablate whether MSE vs Smooth-L1 has an effect on the evaluation criteria:

\begin{equation}
    L_{\text{hyb-mse}}(\mathbf{x}, \mathbf{y}) = \beta \cdot L_{\text{cos}}(\mathbf{x}, \mathbf{y}) + (1-\beta) \cdot L_{\text{mse}}(\mathbf{x}, \mathbf{y})
    \label{eq:hyb_mse}
\end{equation}

In AM-RADIO, the authors used $\beta$ to interpolate between cosine and smooth-L1 loss. Instead of searching the space for the optimal $\beta$, we analyzed the setting they chose ($\beta=0.9$), and also note that cosine loss corresponds to $\beta=1.0$ and MSE loss corresponds to $\beta=0.0$, thus we implicitly study the extremal points of this function interpolation.

\begin{table*}[!t]
\centering\resizebox{\linewidth}{!}{
    \begin{tabular}{lcccccccccc}
        \toprule
        \multirow{2}{*}{\textbf{Model}}                 & \B{Params} & \multicolumn{2}{|c|}{\B{ImageNet1K}} & \multicolumn{2}{c|}{\B{Segmentation (linear)}}   & \multicolumn{4}{|c|}{\B{Vision-Language (LLaVa-1.5)}} & \B{SAM} \\ 
                                                        &  (M)       & \multicolumn{1}{|c}{Zero-shot} 
                                                                                                & \multicolumn{1}{c|}{k-NN}  
                                                                                                              & ADE20k     & \multicolumn{1}{c|}{VOC}   
                                                                                                                                                                        & GQA       & POPE      & TextVQA     & \multicolumn{1}{c|}{VQAv2} 
                                                                                                                                                                                                                          & COCO \\
        \midrule
        \midrule
        AM-RADIO (-H)                           & 653            & \B{82.93}               & \B{86.06} & 51.34  & 84.71 & 63.01 & 86.20     & 56.32       & 79.28 & \B{76.23} \\
        \hline
        PHI-S-RADIO-B                            & 98             & 73.61                    & 81.74      & 48.94  & 84.35 & \ul{63.49} & \ul{86.82}     & \ul{57.64}       & \ul{79.33} & 73.87 \\
        PHI-S-RADIO-L                            & 320            & 81.01                    & 84.68      & \B{51.47} & \B{85.49} & \B{64.29} & \B{86.86} & \B{62.48} & \B{81.10} & 75.06 \\
        \bottomrule
    \end{tabular}
    }
    \caption{Using the PHI Standardization (PHI-S) technique to balance the losses for all of the teachers, we are able to produce ViT-B/16 and ViT-L/16 models using the 3-stage training protocol in appendix \ref{apdx:implementation_details} that are competitive with AM-RADIO (ViT-H/16). Notably, our PHI-S-RADIO-L model achieves higher semantic segmentation results, and significantly higher LLaVA-1.5~\cite{liu2023improvedllava} results. SAM COCO measures the instance mIoU as introduced in \cite{cai2023efficientvit}.}
    \label{tab:headline_metrics}
\end{table*}

\subsection{Normalization}

Instead of balancing the different heads through loss weighting, we can alter the targets themselves. In \cite{wei2022contrastive}, the authors explore this to condition their single teacher's distribution, however, they use the non-invertible LayerNorm operator to rescale the teacher features. Because we want to maintain compatibility for the student to replace the teacher in downstream tasks (by replacing only the vision encoder part of the model), we require the student to still estimate the true teacher distribution. To achieve this, during training, we use an invertible linear mapping $f_k(\cdot)$ such that $T_k'(x) = f_k\left(T_k(x)\right)$ and $T_k(x) = f_k^{-1}\left(T_k'(x)\right)$, where the student model learns to match teacher ($T_k'(x)$) for each of the $k$ teachers.

\subsubsection{Standardization}\label{sec:standardization}

We first consider the simplest case of standardization, which is to use a single scalar $\mu_g$ and std. dev. $\sigma_g$ across the entire feature map. These represent the global statistics of the teacher distribution. In contrast to \cite{wei2022contrastive}, we seek an invertible linear mapping, which excludes LayerNorm. We can, however, estimate the $\mu_{xy}$ and $\sigma_{xy}$ of each position, or, because we want to preserve resolution flexibility, estimate them across all positions and channels, yielding global $\mu_g$ and $\sigma_g$.

\label{sec:global_standardize} Let $\mu_g$ and $\sigma_g$ be the global mean and standard deviation estimate of the teacher distribution. Then

\begin{equation}
    L_\text{gs}(\mathbf{x}, \mathbf{y}) = L_\text{mse}\left(\mathbf{x}, \frac{\mathbf{y} - \mu_g}{\sigma_g} \right)
    \label{eq:global_standardize}
\end{equation}

which we call Global Standardization. \label{sec:standardize} We also explore regular multivariate standardization where we normalize each channel of the teacher distribution independently. Let $\mathbf{\mu}_c = \mathbb{E}\left[\mathbf{Y}_c\right]$ and $\mathbf{\sigma}_c = \sqrt{\text{Var}\left[\mathbf{Y}_c\right]}$, then standardization is defined as

\begin{equation}
\begin{aligned}
    L_s\left(\mathbf{x}, \mathbf{y}\right) = L_\text{mse}(\mathbf{x}, \mathbf{y'}), \quad & y_c' = \frac{y_c - \mu_c}{\sigma_c} 
    \label{eq:standardization}
\end{aligned}
\end{equation}

\subsubsection{Whitening}\label{sec:whitening}\label{sec:zca_whiten}

While standardization normalizes the individual feature variances, it doesn't correct for any covariance between dimensions. We can expand on standardization by also eliminating the covariance between features, called whitening. Let $\cov{Y}$ be the covariance matrix for $\mathbf{Y}$ where $\mathbf{y} \sim \mathbf{Y}$. Following \cite{kessy2018whitening}, we want to find the $\mathbf{W}$ in

\begin{align}
    \mathbf{z} = \mathbf{W}\mathbf{y}
\end{align}

with $\mathbf{z} \sim \mathbf{Z}$ and $\cov{\mathbf{Z}} = \mathbf{I}$. $\mathbf{W} = \cov{Y}^{-\frac{1}{2}}$ is one such valid matrix, called ZCA Whitening \cite{Bell1997THEI}, and takes the form

\begin{align}
    \mathbf{y}' = \cov{Y}^{-\frac{1}{2}} (\mathbf{y} - \bm{\mu})
    \label{eq:zca_whiten}
\end{align}

Each feature in $\cov{Y'}$ is linearly independent and has uniform scale. And so $L_w(\mathbf{x}, \mathbf{y}) = L_{\text{mse}}\left(\mathbf{x}, \mathbf{Wy} - \bm{\mu}\right)$ for any whitening method $w$. $\mathbf{y}$ and $\mathbf{y}'$ are related to each other as

\begin{align}
    \mathbf{y} = \cov{Y}^{\frac{1}{2}}\mathbf{y'} + \bm{\mu}
    \label{eq:zca_whiten_correction}
\end{align}

\subsubsection{Estimation Errors}\label{sec:estimation_errors}
Following the whitening notation of \cite{kessy2018whitening}, given some orthogonal matrix $\mathbf{Q}$, then $\mathbf{QW}$ is also a valid whitening matrix, as $\mathbf{Q}^\intercal \mathbf{Q} = \mathbf{I}$, therefore $\left(\mathbf{QW}\right)^\intercal \mathbf{QW} = \cov{Y}^{-1}$. \cite{kessy2018whitening} then demonstrate the properties of certain choices of $\mathbf{Q}$, and we focus on PCA Whitening (PCA-W) and ZCA in this paper. With

\begin{align}
    \cov{\mathbf{Y}} = \mathbf{U\Lambda U}^\intercal \\
    \mathbf{W}_\text{pca-w} = \mathbf{Q}_{\text{pca-w}}\mathbf{\Lambda}^{-\frac{1}{2}}\mathbf{U}^\intercal, \quad & \mathbf{Q}_\text{pca-w} = \mathbf{I} \label{eq:q_pca_whiten} \\
    \mathbf{W}_\text{zca} = \mathbf{Q}_\text{zca}\mathbf{\Lambda}^{-\frac{1}{2}}\mathbf{U}^\intercal, \quad & \mathbf{Q}_\text{zca} = \mathbf{U} \label{eq:q_zca_whiten}
\end{align}

where $\mathbf{U}$ and $\mathbf{\Lambda}$ are the eigenvectors and eigenvalues for the covariance matrix of $\mathbf{Y}$ respectively. $\bm{\Lambda} = \diagembed\left(\lambda_1,...,\lambda_C\right)$ where $\diagembed\left(\cdot\right)$ forms a diagonal matrix with the vector argument along the diagonal. From \eqref{eq:zca_whiten_correction}, an issue naturally arises, which is the estimation error of our student network. Let $\bm{\epsilon} \in \mathbb{R}^C$ be the estimation error of the student s.t. $\mathbf{y}' = \mathbf{x} + \bm{\epsilon}$ where $\mathbf{x}$ is the student prediction for a given normalized teacher, forming the exact equality

\begin{align}
    \mathbf{y} &= \mathbf{W}^{-1}\left(\mathbf{x} + \bm{\epsilon}\right) + \bm{\mu} \\
               &=  \mathbf{W}^{-1}\mathbf{x} + \mathbf{W}^{-1}\bm{\epsilon} + \bm{\mu} \\
    \bm{\epsilon}_{\text{pca-w}} &= \mathbf{U\Lambda}^{\frac{1}{2}} \bm{\epsilon} \label{eq:error_pca} \\
    \bm{\epsilon}_{\text{zca}} &= \mathbf{U\Lambda}^{\frac{1}{2}}\mathbf{U}^\intercal \bm{\epsilon} \label{eq:error_zca}
\end{align}

We can also use the same $\bm{\epsilon}$ to study standardization (\eqref{eq:standardization}), taking the form

\begin{equation}
    \bm{\epsilon}_{\text{std}} = \diagembed\left(\sigma_1,...,\sigma_C\right)\bm{\epsilon}
    \label{eq:error_stdze}
\end{equation}

As is clear from equations \ref{eq:error_pca}, \ref{eq:error_zca} and \ref{eq:error_stdze}, the choice of normalization will have an impact on the error profile of the model, unless $\bm{\epsilon}$ counteracts the distortion. We next introduce another $\mathbf{Q}$ not studied in \cite{kessy2018whitening}, which is to use a scaled Hadamard matrix, based on this idea.

\subsubsection{Hadamard Whitening (HCA)}\label{sec:hadamard_whiten}

In PCA Whitening, each successive dimension explains the next-largest variance in the data. While this can be a very useful form, we hypothesize that this sort of dimensional loading might not be healthy for a model to learn to match, as effects such as regularization, step size, gradient clipping, etc. may impact the ability of the model to learn each dimension. Instead of ranking the dimensions, we'd like to do the opposite, and find a $\mathbf{Q}$ that explains exactly the same amount of variance irrespective of channel index. It follows that if we could construct an orthogonal basis where each axis captures an identical amount of energy from the diagonal $\mathbf{\Lambda}^{-\frac{1}{2}}$ matrix, then we are able to achieve this balance. First, this matrix $\mathbf{R}$ must be orthogonal for it to be a valid $\mathbf{Q}$. Second, in order for the same proportion of the diagonal $\mathbf{\Lambda}$ to be captured by each row, then each cell must have the same magnitude. Specifically, $\mathbf{R}_{ij} = \pm \frac{1}{\sqrt{C}}$. These matrices are called Hadamard matrices, and the following is called Sylvester's construction \cite{Sylvester1867LXTO}, valid when $C$ is a power of 2:

\begin{equation}    
\begin{aligned}
    \mathbf{H}_1 = \begin{bmatrix}
        1
    \end{bmatrix}, \quad &
    \mathbf{H}_n = \frac{1}{\sqrt{2}} \begin{bmatrix}
        \mathbf{H}_{n-1} & \mathbf{H}_{n-1} \\
        \mathbf{H}_{n-1} & -\mathbf{H}_{n-1}
    \end{bmatrix}
\end{aligned}
\label{eq:sylvester}
\end{equation}

where $n = \log_2 C + 1$. The only difference from standard Sylvester's construction is the $\frac{1}{\sqrt{2}}$ scaling at each recursive level, which is necessary for all of the vectors to be unit length. Relating back to whitening, we use $\mathbf{H}$ as the rotation matrix $\mathbf{Q}$:

\begin{equation}
\begin{aligned}
    \mathbf{W}_\text{hca} = \mathbf{Q}_\text{hca}\mathbf{\Lambda}^{-\frac{1}{2}} \mathbf{U}^\intercal, \quad
    \mathbf{Q}_\text{hca} = \mathbf{H}
\end{aligned}
\label{eq:hadamard_whitening}
\end{equation}

and we end up with ``Hadamard Whitening'' with corresponding error profile:

\begin{equation}
    \bm{\epsilon}_{\text{hada}} = \mathbf{U\Lambda}^{\frac{1}{2}}\mathbf{H}^\intercal \epsilon
    \label{eq:error_hadamard}
\end{equation}

This error profile is interesting due to the fact that an error of size $\delta$ along any single dimension $d_1$ will have identical magnitude in the original space as any other dimension $d_2$. We prove this in appendix \ref{sec:hca_error_profile}. Further, in appendix \ref{sec:general_hadamard_matrices} we show how some Hadamard matrices whose size is not a power of 2 can be constructed, and how we found an $\mathbf{H}$ for important model sizes such as $768$, $1024$, $1152$, $1280$, and $1408$.

\subsubsection{PCA-Hadamard Isotropic Standardization (PHI-S)}\label{sec:ship}

A key issue with the previous normalization procedures (aside from global standardization) is that they place disproportionate weight on lower-variance axes. To avoid this distortion, we present the following theorem, and then describe how we apply it as a novel form of standardization:

\begin{theorem}
    For any mean-centered normal data distribution $\mathbf{X} \in \mathbb{R}^{C \times N}$ with satisfiable Hadamard-matrix dimension $C$, there exists an orthogonal transform $\mathbf{R} \in \mathbb{R}^{C \times C}$ and scalar $\alpha \in \mathbb{R}$ such that $\textup{diag}\left(\cov{\alpha\mathbf{RX}}\right) = \mathbf{1}_C$.
\end{theorem}

\begin{proof}
    Let $\cov{\mathbf{X}}$ be the covariance matrix of $\mathbf{X}$, and let $\cov{\mathbf{X}} = \mathbf{U\Lambda U}^\intercal$ where $\mathbf{U}$ is an orthogonal matrix, and $\mathbf{\Lambda} = \diagembed\left(\lambda_1,...,\lambda_C\right)$, with $\lambda_i$ being the eigenvalues of $\cov{\mathbf{X}}$. (called PCA).

    First, note that $\cov{\mathbf{U}^\intercal \mathbf{X}} = \mathbf{U}^\intercal \left(\mathbf{U\Lambda U}^\intercal\right) \mathbf{U} = \mathbf{\Lambda}$.

    Next, let $\mathbf{H} \in \mathbb{R}^{C \times C}$ be a normalized Hadamard matrix, and recall each cell in $\mathbf{H}$ has value $\pm \frac{1}{\sqrt{C}}$. Using the orthogonal transform $\mathbf{HU}^\intercal$, we get $\cov{\mathbf{HU}^\intercal \mathbf{X}} = \mathbf{H\Lambda H}^\intercal$.

    \begin{align}
        \diag\left(\mathbf{H\Lambda H}^\intercal\right)_r &= \sum_{i=1}^C \lambda_i \left(\pm \frac{1}{\sqrt{C}}\right)^2 = \frac{1}{C} \sum_i^C \lambda_i \quad \forall r \in C
    \end{align}

    Let 
    \begin{align}
        \phi &= \sqrt{\frac{1}{C} \sum_i^C \lambda_i} \label{eq:ship_phi} \\
        \cov{\phi^{-1} \mathbf{M}} &= \phi^{-2} \mathbf{M} = \frac{C}{\sum_i^C \lambda_i} \mathbf{M}
    \end{align}

    for some matrix $\mathbf{M}$. For $\mathbf{H\Lambda H}^\intercal$, we have
    
    \begin{align}
        \diag\left(\cov{\phi^{-1} HU^\intercal X}\right)_r = \diag\left(\phi^{-2} \mathbf{H\Lambda H}^\intercal\right)_r &= \frac{C \sum_i^C \lambda_i}{C \sum_i^C \lambda_i} = 1 \quad \forall r \in C
    \end{align}
    
    Therefore 

    \begin{equation}
    \begin{aligned}
        \mathbf{R} = \mathbf{HU}^\intercal \quad & \alpha = \phi^{-1}
        \label{eq:ship_proof}
    \end{aligned}
    \end{equation}

\end{proof}

For PHI-S, following \eqref{eq:ship_proof} we use

\begin{equation}
    \mathbf{W}_\text{ship} = \alpha \mathbf{R}
    \label{eq:ship}
\end{equation}

Essentially, we first mean center and then rotate the distribution in such a way ($\mathbf{R}$) that the variance along each resulting dimension is identical, allowing us to uniformly scale by $\alpha$ to achieve a standardized distribution.

\begin{figure*}[!ht]
  \centering\resizebox{\linewidth}{!}{
    \includegraphics{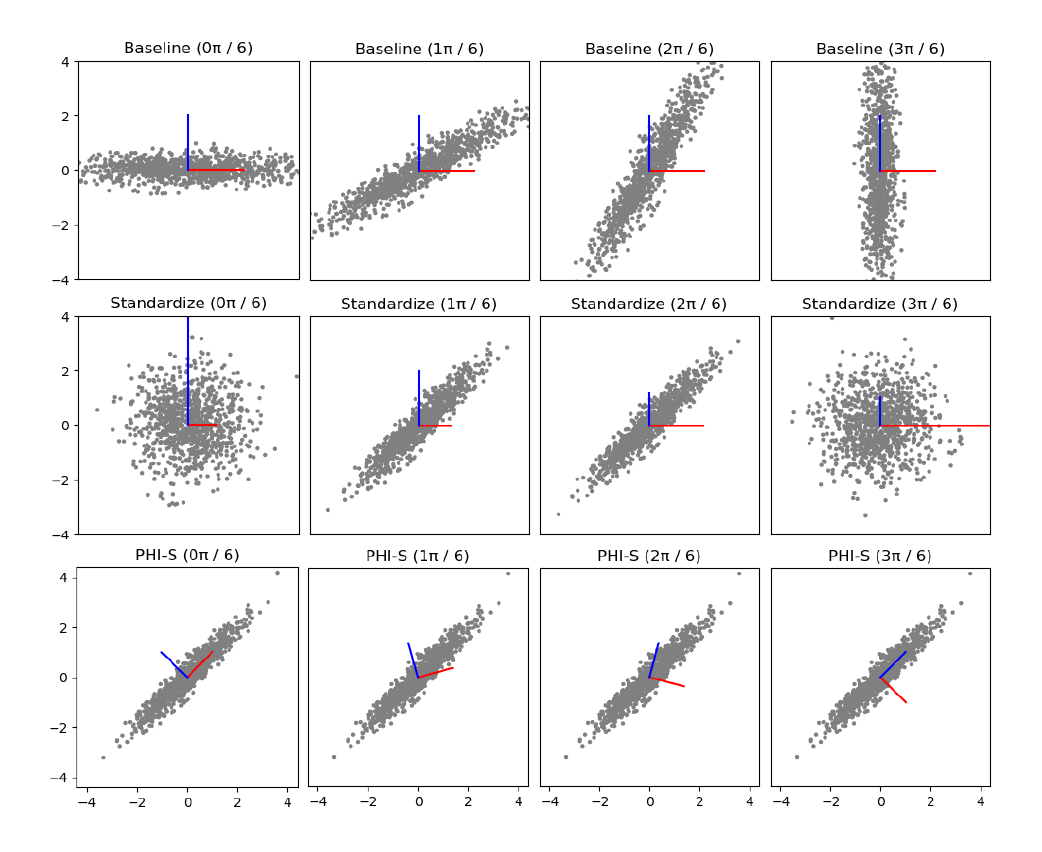}
  }
  
  \caption{
  Visualization of how standardization affects the resulting data distribution. We start with the same distribution, and rotate the data by some angle. Regular standardization's effect is directly tied to the distribution orientation. Conversely, PHI-S is invariant to any data rotation, and will produce an identical transform up to sign along each dimension. We can make the sign consistent by negating the rows of $\mathbf{H}$ and $\mathbf{U}$ which have a negative value in the diagonal position. Similarly, regular standardization will distort each dimension (shown with red/blue lines), which will have the effect of reducing the importance of high variance axis-aligned dimensions, and increasing the importance of low-variance dimensions. PHI-S is isotropic, so the change in scale is uniform.
  }
  \label{fig:ship_viz}
\end{figure*}

\subsubsection{Visualizing Distributions}

\begin{figure*}[!ht]
  \centering\resizebox{\linewidth}{!}{
    \includegraphics{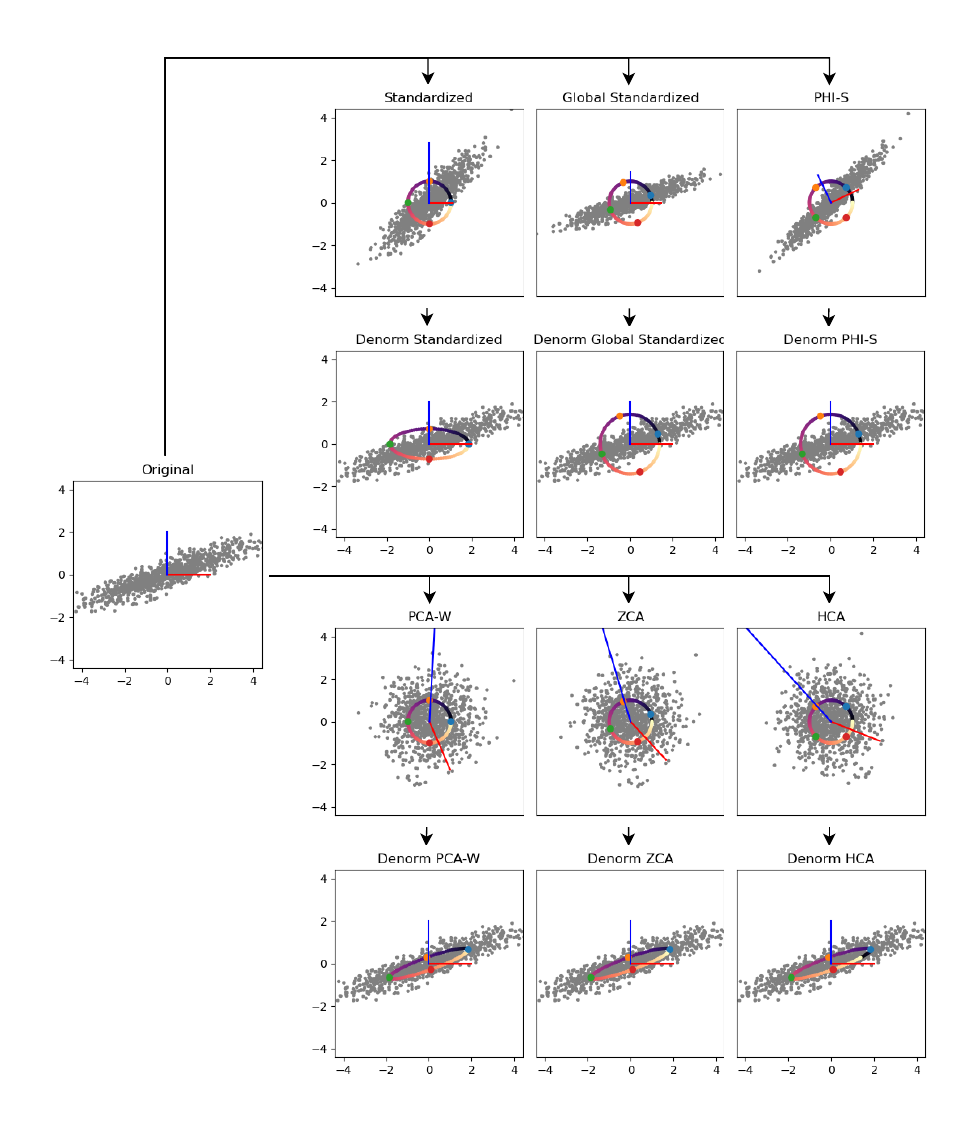}
  }
  
  \caption{
  Visualization of normalization procedures. We display two axis lines in red and blue. In the original space, they're both 2 units long, and aligned with the plot coordinate system. We also display an ``error circle'' which is a unit circle in the normalized coordinate system. For the three whitening transforms you can see how they only differ by rotation. We also specifically draw colored dots on the error circle corresponding to the extremal points of the error circle when denormalized into an ellipse. PCA-W places the largest error magnitude on the x-axis, given that it's the dimension with largest eigenvalue thus estimation errors along the x dimension will have a much larger impact in the denormalized space. As we show in \eqref{eq:error_zca}, the error for ZCA will be proportional to the original distribution's covariance matrix, and thus, the extremal points are along the eigenvectors of the covariance matrix. Hadamard whitening has the extremal points at $\left|x_1\right| = \left|x_2\right| = ... = \left|x_C\right|$. Global Standardization and PHI-S are both isotropic, which means that there's an infinite number of extremal points, so we instead show the points as they relate to the distribution itself. Similar to ZCA, for Global Standardization these points are along the principal axes. And similar to HCA, the aligned points for PHI-S are when $\left|x_1\right| = \left|x_2\right| = ... = \left|x_C\right|$.
  }
  \label{fig:transform_viz}
\end{figure*}

In figure \ref{fig:transform_viz} we show how the various normalization transforms change the target distribution, and also how the transforms affect the errors coming from the student model. For the whitening transforms, the choice of $\mathbf{Q}$ matrix has an impact on the relationship between errors of the same magnitude (e.g. fixed radius) in the learned distribution versus the denormalized distribution. Using the Hadamard matrix as $\mathbf{Q}$ is the only choice that doesn't place extra error on a particular learned dimension.

\begin{figure*}[!ht]
    \centering\resizebox{0.7\linewidth}{!}{
        \includegraphics{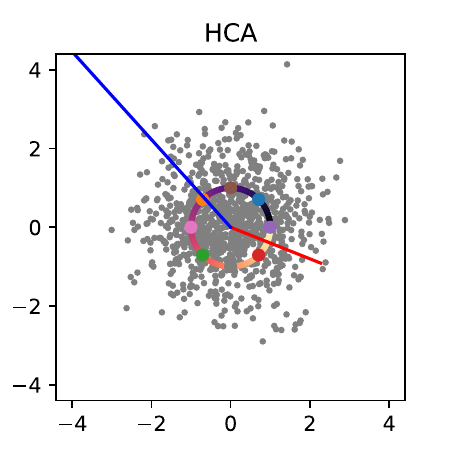}
        \includegraphics{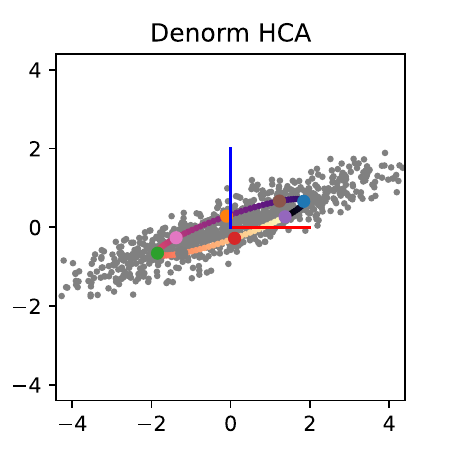}
    }
    \caption{
        Related to figure \ref{fig:transform_viz} and \eqref{eq:hadamard_delta_error}, we visualize what happens to the one-hot error vectors when projecting back to the original space for HCA. We retain the original $|x|=|y|$ dots, and add the one-hot dots demonstrating how their mapping remains equidistant from the origin relative to each other. In particular, since $\delta = 1$, then $\left\lVert \left(\mathbf{U\Lambda}^\frac{1}{2}\mathbf{H}^\intercal\right) \mathbf{\Delta}_r \right\rVert = \sqrt{\frac{1}{C} \sum_c \lambda_c} \approx 1.400892$ for any choice of $r$. For reference, $\mathbf{\Lambda} \approx \left[3.8356, 0.0894\right]$.
    }
    \label{fig:hca_onehot}
\end{figure*}

\begin{figure*}[!ht]
    \centering\resizebox{0.7\linewidth}{!}{
        \includegraphics{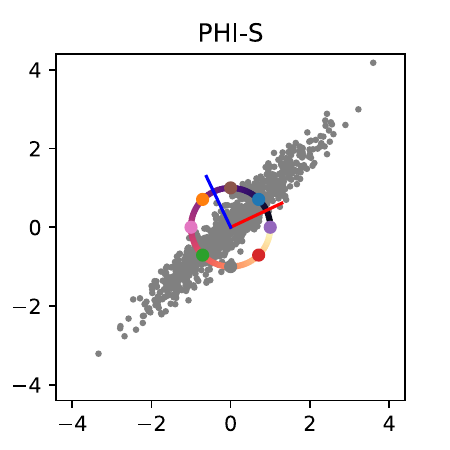}
        \includegraphics{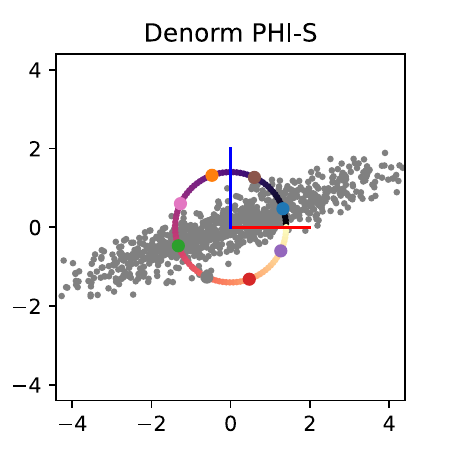}
    }
    \caption{
        Similar to figure \ref{fig:hca_onehot} and figure \ref{fig:transform_viz}, we visualize PHI-S in the normalized and denormalized spaces. This visualizes how \eqref{eq:error_ship} maintains errors along a circle in both spaces, owing to the isotropic nature of the transform. It also can be seen how the $|x|=|y|$ error dots in normalized space map to the principal directions of the distribution, and also how the one-hot dots capture identical probability density.
    }
    \label{fig:ship_onehot}
\end{figure*}

\begin{figure*}[!ht]
    \centering\resizebox{\linewidth}{!}{
        \includegraphics{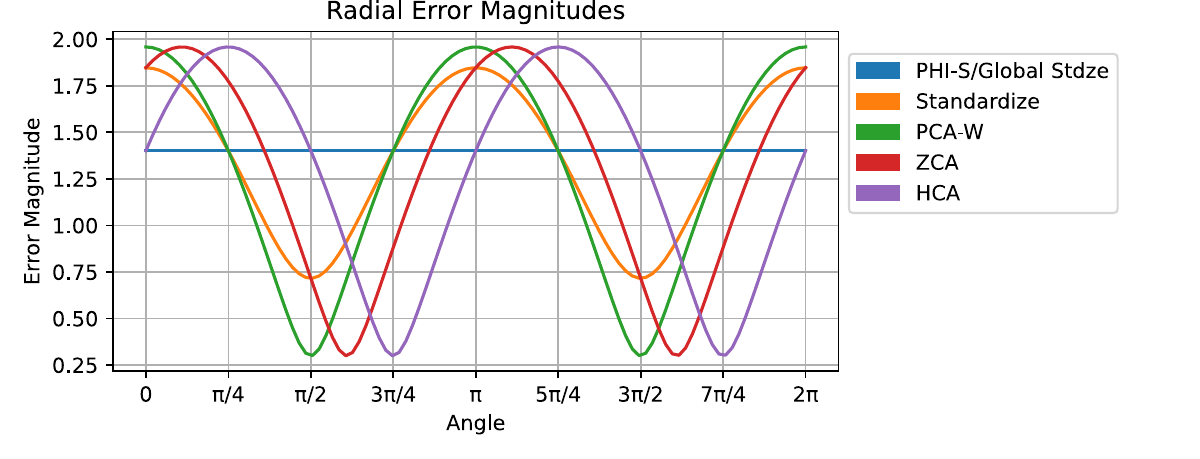}
    }
    \caption{
        Following from figure \ref{fig:transform_viz}, we visualize the radius of the denormalized error circle at every angle between 0 and $2\pi$. Because Global Standardization and PHI-S are isotropic, and because the distribution is mean centered (section \ref{sec:gstd_vs_phis}), they scale the error circle uniformly by the same amount. As predicted, for $\theta = z\frac{\pi}{2}$ with $z \in \mathbb{Z}$ (e.g. when $y = 0$ or $x = 0$) we have the same error magnitude for HCA, and also where PHI-S and HCA have identical magnitude. HCA has extremal values at $\theta_{\text{hca}}^{\text{ex}} = z\frac{\pi}{2} + \frac{\pi}{4}$. PCA-W has extremal values at $\theta_{\text{pca-w}}^{\text{ex}} = z\frac{\pi}{2}$. We also have that ZCA will have extremal values $\theta_{\text{pca-w}}^{\text{ex}}(z) \leq \theta_{\text{zca}}^{\text{ex}}(z) \leq \theta_{\text{hca}}^{\text{ex}}(z)$.
    }
    \label{fig:radial_error_magnitudes}
\end{figure*}

In figure \ref{fig:radial_error_magnitudes} we display the radius of the denormalized error circle. An interesting property of standardization becomes apparent, which is that the error magnitude of standardization is bounded between PCA-W and PHI-S, with equality at $\cov{Y} = \bm{\Lambda}$ for the former and $\diag\left(\cov{Y}\right) = \phi_{\text{ship}}\mathbf{I}$ for the latter. One hypothesis for why the standardization transforms (Global Standardization, Standardization, PHI-S) work best in our study is because the error amplitudes are ``less extreme'' than whitening in general. With MSE being sensitive to outliers, this property is likely important. Because the whitening methods only differ by an orthogonal transform, their errors are phase shifted relative to each other. 

\section{Implementation Details}\label{sec:ablation_impl_details}

We generally follow the procedure outlined in AM-RADIO, however we make some changes that reduce the computational cost of training, which was necessary to cover all of the ablations we studied. Namely, we:
\begin{itemize}
    \item Add SigLIP as a teacher.
    \item Train the student model at 256px resolution, and downsample the teacher features to match.
    \item Train for 300k steps instead of the 600k steps originally proposed.
    \item Split each teacher into their own partition, resulting in each teacher receiving a batch of 256 images, with a total of 1024 images per iteration.
    \item Initialize from TIMM \cite{rw2019timm} ``vit\_[base,large]\_patch16\_224'' pretrained models.
\end{itemize}

We found that downsampling SAM features degrades their quality, so instead we pad the small image and crop out the features. Further details, and specifically for table \ref{tab:headline_metrics}, are presented in appendix \ref{apdx:implementation_details}.

\section{Results}

In figure \ref{fig:loss_plots} we display our model's ability to estimate the teacher distributions during training. For any of the transforms that project the teacher features into a different space, we apply the inverse operation so that all methods are measured in the original space. As can be seen, ``Baseline'' is much worse than any other method, and it's intuitive because it allows the relative difference in magnitudes between the different teachers to implicitly weight the loss. SAM has much larger activation variance than any other model, which results in the Baseline model spending most of its energy trying to match SAM. Overall, the PHI Standardization method produces the best results, as it's able to simultaneously beat any other method on DFN CLIP, SigLIP, second best on DINOv2, while remaining competitive on SAM. We show the final MSEs in table \ref{tab:raw_mse}.

\begin{table}[]
    \centering
    \resizebox{0.65\linewidth}{!}{
    \begin{tabular}{r|cccc}
        \B{Method} $\downarrow$   & \B{DFN CLIP} ($\cdot 1^{-4}$)    & \B{SigLIP} & \B{DINOv2} & \B{SAM}    \\
        \hline
        MSE           & 5.0883           & 1.9598      & 1.0767      & \B{6.5082}  \\
        Cosine        & 105.90           & 3.3060      & 1.7980      & 27.9310     \\
        Hyb MSE       & 7.4930           & 1.9250      & 0.9422      & \ul{7.4580} \\
        Hyb SmL1      & 9.8540           & 1.9750      & 0.9112      & 8.6600      \\
        \hline
        Global Stdze  & 4.7420           & \ul{1.9120} & \B{0.8801} & 8.4910       \\
        Standardize   & \ul{4.7417}      & 1.9146      & 0.8928      & 8.3272      \\
        \B{PHI-S (Ours)}   & \B{4.7200}      & \B{1.9010} & \ul{0.8865} & 8.3330        \\
        \hline                                                                     
        PCA-W         & 4.7861           & 1.9534      & 0.9316      & 8.7309      \\
        ZCA           & 4.7841           & 1.9529      & 0.9321      & 8.7061      \\
        HCA (Ours)    & 4.7855           & 1.9545      & 0.9326      & 8.7226      \\
    \end{tabular}
    }
    \caption{Mean Squared Error for matching the teachers with a ViT-B/16 student using different algorithms. PHI-S does the best job at simultaneously minimizing all teachers.}
    \label{tab:raw_mse}
\end{table}

\begin{table*}[]
    \centering
    \resizebox{\linewidth}{!}{
    \begin{tabular}{r|cccccc|cc}
                         & \B{Feature}     & \multirow{2}{*}{\B{Classification}} & \multirow{2}{*}{\B{Segmentation}} & \B{SAM}    & \B{LLaVA}   & \B{Probe}   & \multirow{2}{*}{\B{Average}} & \B{Average} \\
                         & \B{MSE}         &                &              & \B{COCO}   & \B{1.5}     & \B{3D}      &           & \B{No COCO}           \\
            \hline
        Baseline MSE & 3.25            & 4.00           & 4.00         & \B{1.00}   & 4.00        & 4.00        & 3.38      & 3.85      \\
        Global Stdze & 2.75            & \ul{2.50}      & 2.50         & 3.00       & \B{1.875}   & 2.25        & 2.48      & 2.38      \\
        Standardize  & \ul{2.25}       & \ul{2.50}      & \ul{2.00}    & \ul{2.00}  & 2.25        & \B{1.75}    & \ul{2.13} & \ul{2.15} \\
        \B{PHI-S}         & \B{1.75}        & \B{1.00}       & \B{1.50}     & 4.00       & \B{1.875}   & \B{1.75}    & \B{1.98}  & \B{1.58}  \\
    \end{tabular}
    }
    \caption{Average benchmark ranks for the ViT-L/16 models using the best (and baseline) normalization methods from the ViT-B/16 ablations. PHI-S is even more dominant with the larger model. We provide the raw benchmark scores in appendix \ref{sec:vitl_raw_scores}.}
    \label{tab:vitl_method_ranks}
\end{table*}

In tables \ref{tab:vitb_method_ranks} and \ref{tab:vitl_method_ranks}, we display the average benchmark ranks across different benchmarks and methods for ViT-B/16 and ViT-L/16 students, respectively. For LLaVA, we first average the two GQA and TextVQA tasks separately, and then combine them with POPE and VQAv2 to compute the average. This is to prevent overly biasing towards the tasks that have multiple measurements. In both architectures PHI-S produces the best results by achieving the lowest average rank across the suite.

\subsection{Empirical Errors}

In section \ref{sec:estimation_errors} we demonstrated how the choice of normalization might have an impact on the errors the student makes when matching the teachers. Particularly, \eqref{eq:error_zca} is the error profile for ZCA, \ref{eq:error_pca} for PCA-W, \ref{eq:error_hadamard} for HCA, and \ref{eq:error_stdze} for regular standardization. We also have

\begin{equation}
\begin{aligned}
    \bm{\epsilon}_{gs} = \alpha_{gs}^{-1} \bm{\epsilon} \quad\quad &
    \bm{\epsilon}_{\text{ship}} = \alpha_{\text{ship}}^{-1} \bm{\epsilon}
\end{aligned}
    \label{eq:error_ship}
\end{equation}

for global standardization and PHI-S respectively. We used this error profile to motivate the introduction of Hadamard matrices for whitening in section \ref{sec:hadamard_whiten}, as it distributes the error variance equally through all channels of the denormalization projection. In table \ref{tab:denorm_error_ranges} we display the empirical error variance ranges for each studied method and for each teacher. Intriguigingly, both methods that employ the Hadamard matrix (HCA and PHI-S) have very low variance ranges compared to the other methods. This implies that the student model is making errors of roughly uniform magnitude across all channels. Unfortunately, in the case of HCA, this property isn't borne out in a useful way in the benchmarks (table \ref{tab:vitb_method_ranks}). Table \ref{tab:denorm_error_ranges} shows that the loss landscape and/or the optimizer are adapting to normalization distortions and baking the non-uniform nature of the variances into the student model. For PHI-S, the student model still has nearly uniform error variance in the normalized space, but also has the lowest (or nearly lowest) range in the denormalized (original) space. This isn't surprising given that a unit change in any dimension of the normalized space has an identical effect as any other dimension, thus there's no incentive to prefer one dimension to another.

\begin{figure*}[!ht]
  \centering\resizebox{\linewidth}{!}{
    \includegraphics{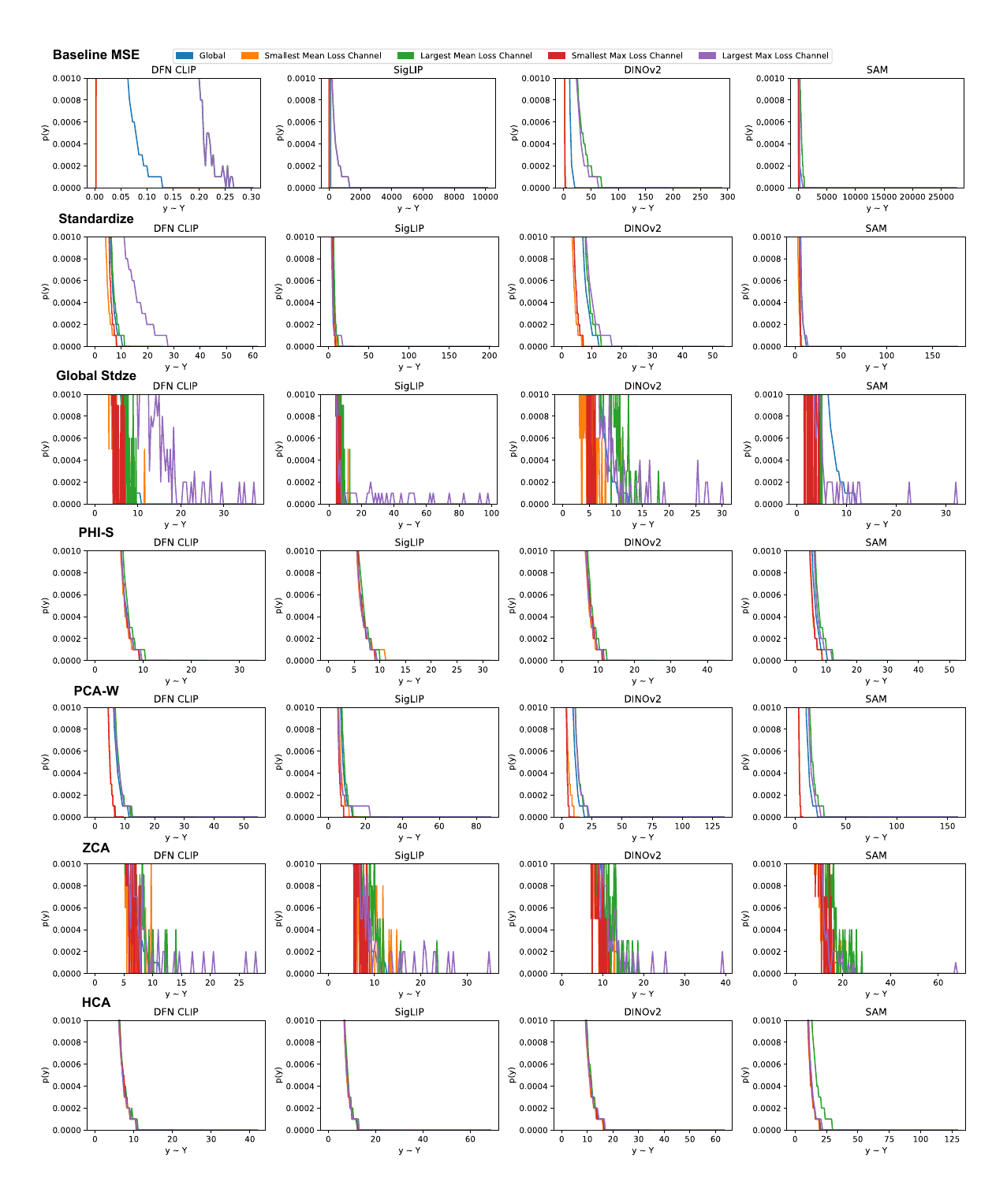}
  }
  
  \caption{
  Loss distributions for various normalization methods. The x-axis range is based on the minimum and maximum losses seen for each method over the course of 1,000 samples after training for 100k iterations. The ``Largest Max Loss Channel'' shows the distribution for the channel that had the highest loss value. It helps us understand how vulnerable our learning process is to outliers. The ``Global'' curve shows the distribution by combining all of the channels.
  }
  \label{fig:loss_histograms}
\end{figure*}

\begin{table*}[!ht]
    \centering
    \resizebox{\linewidth}{!}{
    \begin{tabular}{r|cccc|cccc}
    \multirow{2}{*}{\B{Method}} & \multicolumn{4}{c}{\B{Normalized}} & \multicolumn{4}{c}{\B{Denormalized}} \\
                                & \B{DFN CLIP} & \B{SigLIP} & \B{DINOv2} & \B{SAM}    & \B{DFN CLIP} & \B{SigLIP}   & \B{DINOv2} & \B{SAM} \\
    \hline
    Baseline - MSE          & 0.015        & 403.995    & 2.705      & 238.929    & 0.015        & 403.995      & 2.705      & \B{238.929} \\
    \hline
    Global Stdze            & 17.759       & 113.783    & 1.287      & 8.594      & \B{0.014}    & \ul{398.198} & \B{2.405}  & 239.744 \\
    Standardize             & 0.579        & 0.348      & 0.480      & 0.861      & 0.015        & 406.442      & 2.793      & 240.526 \\
    PHI-S                    & \ul{0.086}   & \ul{0.088} & \ul{0.052} &\B{0.219}   & \B{0.014}    & \B{393.216}  & \ul{2.447} & \ul{239.489} \\
    \hline
    PCA-W                   & 0.416        & 0.339      & 0.830      & 1.634      & 0.015        & 421.195      & 3.179      & 243.610 \\
    ZCA                     & 0.558        & 0.368      & 0.626      & 1.226      & 0.015        & 421.192      & 3.098      & 243.774 \\
    HCA                     & \B{0.028}    & \B{0.030}  & \B{0.035}  & \ul{0.232} & 0.015        & 422.810      & 3.137      & 243.596
    
    \end{tabular}
    }
    \caption{The \textit{range} of the per-channel variances of both the normalized student model errors, as well as the denormalized student errors. A smaller range implies that each channel has a more similar error variance, with $0$ implying that each channel has identical error variance. As theorized, Hadamard and PHI-S have the most uniform variances across the channels, however PHI-S also has the most uniform error variance when projected back into the original (denormalized) space.}
    \label{tab:denorm_error_ranges}
\end{table*}

In figure \ref{fig:loss_histograms} we show the loss distributions for the core normalization methods we studied. It helps us understand not only the magnitudes of the errors, but also showcases how different normalization methods affect the behavior of outliers. It's very apparent that ``Baseline'' has uncontrolled magnitudes, with SAM having quite extreme losses, especially relative to DFN CLIP. This is also where we can really see how ``Global Standardize'' and ``PHI-S'' differ in behavior, owing to PHI-S equializing the variance across channels. The purple curve shows how global standardization is still very susceptible to outlier errors. As predicted in section \ref{sec:estimation_errors}, the methods that use Hadamard matrices (PHI-S and HCA) have the tightest error bounds between channels. Finally, it's also apparent how well PHI-S works for balancing across teachers, as the losses all have the most similar distributions compared against the other methods.

\section{Related Work}
\noindent\paragraph{Knowledge Distillation}
We base our work on \cite{ranzinger2023amradio} which considers a multi-teacher distillation problem without ground-truth labels, and where the targets are the teacher features themselves, instead of estimating e.g. a probability distribution for classification. They build upon extensive literature on knowledge distillation, popularized by \cite{hinton2015distilling}, and then expanded with \cite{kim2018paraphrasing,ba2014deep,Mirzadeh2019ImprovedKD,beyer2022goodteacher} for teacher logit estimation problems. For feature matching, \cite{Romero2014FitNetsHF,huang2017like,ahn2019variational,heo2019overhaul,zagoruyko2017paying,sun2021dynamic,wei2022featuredistillation} study this sub-problem. Specifically, \cite{wei2022contrastive} discuss the importance of normalizing the teacher feature distribution, which is a notable omission in \cite{ranzinger2023amradio}. Further, in the knowledge distillation domain, the idea of distilling from multiple teachers at once is heavily studied \cite{hinton2015distilling,liu2020adamultiteachr,zuchniak2023multiteacher,yuan2020reinforced,zhao2022collabteaching,yang2020modelcompression,Park2020FeatureLevelEK,you2017multiteacher,lan2018knowledge,Asif2019EnsembleKD,Fukuda2017EfficientKD}. AM-RADIO \cite{ranzinger2023amradio} differentiates itself from those largely through the lack of a unified target label or distribution, as the teachers aren't even from the same problem domain (e.g. CLIP \cite{radford2021clip} versus SAM \cite{kirillov2023sam}), and thus will produce very different feature distributions for the same image. Similarly, much of the literature that covers balancing the multi-objective (multi-teacher) loss relies on having access to ground truth labels \cite{liu2020adamultiteachr}. Generically, \cite{hu2019adaloss} is capable of balancing losses without GT labels by setting the loss weight to be inversely proportional to the approximate expected loss for each term, which AM-RADIO studied but found no significant effect. In \cite{ruder2017knowledgeadaptationteachingadapt}, the authors study domain adaptation where they have multiple classifier teachers from their own domain, and they seek to train a student on a new unlabeled domain, however their method relies on the source and target domains being classification. Concurrently to our work, \cite{shang2024theia} introduced the ``Theia'' model which draws heavily from AM-RADIO including the loss formulation. In their work, the authors chose to use the regular standardization method, a choice which this work explores and demonstrates that it was both a great addition over AM-RADIO, but also not the optimal choice compared against PHI-S which we propose here. We view the works as complementary, as our study entirely revolves around the design choices in their section 3.2 and AM-RADIO's section 3.4. Recently, the preprint UNIC~\cite{sariyildiz2024unicuniversalclassificationmodels} is also based on AM-RADIO, and employs feature standardization, showing strong positive effects, and preprint UNIT~\cite{zhu2024unit} bases on AM-RADIO employing feature standardization in addition to explicit supervised OCR learning.

\noindent\paragraph{Normalization}
The importance of normalization in distillation was identified in \cite{heo2019overhaul}, which used BatchNorm. More recently, \cite{wei2022contrastive} also considered normalized feature matching, however their choice of LayerNorm was non-invertible, and also doesn't de-correlate the different feature dimensions. We aim to preserve the ability of the student to estimate the teacher as in AM-RADIO, so we focus on invertible normalization techniques which allow us to estimate the teacher's true distribution. \cite{liu2022normalizedfeaturedistillationsemantic} argue that normalizing the student and teacher features improves distillation for semantic segmentation as the student otherwise spends most of its energy matching the teacher magnitudes. Intuitively, we expand on this by also observing that controlling the relative magnitudes across teachers is critical. \cite{kessy2018whitening} provides an overview of different whitening procedures, stressing the fact that there are infinitely many whitening matrices for a given distribution, and focus their attention on the $\mathbf{Q}$ rotation matrix that relates them. Their treatment covers many popular $\mathbf{Q}$ matrices, and we use their work as the foundation for our study. There are also multiple works in the SSL vision domain that deal with distribution properties, such as Barlow Twins~\cite{zbontar2021barlow} and VICReg~\cite{bardes2022vicreg}. Their algorithms try to induce the model to produce regular features, where in contrast, we're forced to deal with arbitrary models that didn't undergo such regularization. In digital signal processing, using the Hadamard matrix to spread energy (to mitigate signal loss errors) is a common practice \cite{Pratt1969Hadamard,kanj2022whitening}. We study the incorporation of this matrix both as a suitable $\mathbf{Q}$ matrix for rotation during the whitening process, and also in a novel way to derive a scalar normalization factor that standardizes any multivariate distribution with a known Hadamard matrix, which we call PHI Standardization (PHI-S).

\section{Conclusion}

Through our experiments, we have conclusively demonstrated that using plain MSE without balancing has a large negative effect on the resulting quality of the student model. Among normalization methods, standardization worked better than whitening, which was an initially surprising result. We hypothesize that the major issue with whitening is that the teacher models aren't producing full rank distributions (appendix, table \ref{tab:teacher_effective_ranks}), which makes the normalization factors unstable. Regular standardization is resistant to this because the principal components of the distribution are spread out across all of the dimensions, preventing degenerate $\bm{\Lambda}^{-\frac{1}{2}}$ solutions. We found two novel applications of Hadamard matrices with respect to distribution normalization: HCA and PHI-S. At the ViT-B/16 model scale, we found that isotropic normalization methods (Global Standardize and PHI-S) worked the best, and for ViT-L/16, PHI-S remained the best. On the topic of reconstruction errors, we found no significant result across the whitening methods with respect to downstream metrics, and also found that the per-channel estimation errors were not uniform in general, unless uniform is the optimal choice (HCA and PHI-S), implying that the student model is able to be robust to the potentially high-distortion nature of the different transforms. Overall, PHI-S appears to be the best normalization method studied, and it allowed us to produce ViT-B and ViT-L models that are competitive with the original AM-RADIO~\cite{ranzinger2023amradio} ViT-H model. 

\noindent\B{Future Work}
We've solely explored the use of PHI-S for agglomerative modeling, however it's a general standardization technique when certain assumptions about the data hold such as normality and dimensionality of the distribution. PHI-S could additionally be used to post-hoc standardize the output of existing models. Lastly, an opportunity arises when combining PHI-S with quantization practices (similar to \cite{ashkboos2024quarotoutlierfree4bitinference}) in the information retrieval domain as it balances the information across all channels evenly, potentially unlocking higher fidelity quantizers.

\bibliography{phis}

\begin{thebibliography}{62}
\providecommand{\natexlab}[1]{#1}
\providecommand{\url}[1]{\texttt{#1}}
\expandafter\ifx\csname urlstyle\endcsname\relax
  \providecommand{\doi}[1]{doi: #1}\else
  \providecommand{\doi}{doi: \begingroup \urlstyle{rm}\Url}\fi

\bibitem[Ahn et~al.(2019)Ahn, Hu, Damianou, Lawrence, and Dai]{ahn2019variational}
S.~Ahn, S.~Hu, A.~Damianou, N.~D. Lawrence, and Z.~Dai.
\newblock Variational information distillation for knowledge transfer.
\newblock In \emph{2019 IEEE/CVF Conference on Computer Vision and Pattern Recognition (CVPR)}, pp.\  9155--9163, Los Alamitos, CA, USA, jun 2019. IEEE Computer Society.
\newblock \doi{10.1109/CVPR.2019.00938}.
\newblock URL \url{https://doi.ieeecomputersociety.org/10.1109/CVPR.2019.00938}.

\bibitem[Ashkboos et~al.(2024)Ashkboos, Mohtashami, Croci, Li, Jaggi, Alistarh, Hoefler, and Hensman]{ashkboos2024quarotoutlierfree4bitinference}
Saleh Ashkboos, Amirkeivan Mohtashami, Maximilian~L. Croci, Bo~Li, Martin Jaggi, Dan Alistarh, Torsten Hoefler, and James Hensman.
\newblock Quarot: Outlier-free 4-bit inference in rotated llms, 2024.
\newblock URL \url{https://arxiv.org/abs/2404.00456}.

\bibitem[Asif et~al.(2019)Asif, Tang, and Harrer]{Asif2019EnsembleKD}
Umar Asif, Jianbin Tang, and Stefan Harrer.
\newblock Ensemble knowledge distillation for learning improved and efficient networks.
\newblock In \emph{European Conference on Artificial Intelligence}, 2019.
\newblock URL \url{https://api.semanticscholar.org/CorpusID:202660953}.

\bibitem[Assran et~al.(2023)Assran, Duval, Misra, Bojanowski, Vincent, Rabbat, LeCun, and Ballas]{assran2023ijepa}
Mahmoud Assran, Quentin Duval, Ishan Misra, Piotr Bojanowski, Pascal Vincent, Michael Rabbat, Yann LeCun, and Nicolas Ballas.
\newblock Self-supervised learning from images with a joint-embedding predictive architecture, 2023.
\newblock URL \url{https://arxiv.org/abs/2301.08243}.

\bibitem[Awais et~al.(2023)Awais, Naseer, Khan, Anwer, Cholakkal, Shah, Yang, and Khan]{awais2023foundational}
Muhammad Awais, Muzammal Naseer, Salman Khan, Rao~Muhammad Anwer, Hisham Cholakkal, Mubarak Shah, Ming-Hsuan Yang, and Fahad~Shahbaz Khan.
\newblock Foundational models defining a new era in vision: A survey and outlook, 2023.

\bibitem[Ba \& Caruana(2014)Ba and Caruana]{ba2014deep}
Jimmy Ba and Rich Caruana.
\newblock Do deep nets really need to be deep?
\newblock In \emph{Advances in Neural Information Processing Systems}, pp.\  2654--2662, 2014.

\bibitem[Bardes et~al.(2022)Bardes, Ponce, and LeCun]{bardes2022vicreg}
Adrien Bardes, Jean Ponce, and Yann LeCun.
\newblock {VICR}eg: Variance-invariance-covariance regularization for self-supervised learning.
\newblock In \emph{International Conference on Learning Representations}, 2022.
\newblock URL \url{https://openreview.net/forum?id=xm6YD62D1Ub}.

\bibitem[Bell \& Sejnowski(1997)Bell and Sejnowski]{Bell1997THEI}
Anthony~J. Bell and Terrence~J. Sejnowski.
\newblock The ``independent components''' of natural scenes are edge filters 3329 recover the causes.
\newblock 1997.
\newblock URL \url{https://api.semanticscholar.org/CorpusID:18326486}.

\bibitem[Beyer et~al.(2022)Beyer, Zhai, Royer, Markeeva, Anil, and Kolesnikov]{beyer2022goodteacher}
L.~Beyer, X.~Zhai, A.~Royer, L.~Markeeva, R.~Anil, and A.~Kolesnikov.
\newblock Knowledge distillation: A good teacher is patient and consistent.
\newblock In \emph{2022 IEEE/CVF Conference on Computer Vision and Pattern Recognition (CVPR)}, pp.\  10915--10924, Los Alamitos, CA, USA, jun 2022. IEEE Computer Society.
\newblock \doi{10.1109/CVPR52688.2022.01065}.
\newblock URL \url{https://doi.ieeecomputersociety.org/10.1109/CVPR52688.2022.01065}.

\bibitem[Buciluǎ et~al.(2006)Buciluǎ, Caruana, and Niculescu-Mizil]{bucilu2006model}
Cristian Buciluǎ, Rich Caruana, and Alexandru Niculescu-Mizil.
\newblock Model compression.
\newblock In \emph{Proceedings of the 12th ACM SIGKDD international conference on Knowledge discovery and data mining}, pp.\  535--541. ACM, 2006.

\bibitem[Cai et~al.(2023)Cai, Li, Hu, Gan, and Han]{cai2023efficientvit}
Han Cai, Junyan Li, Muyan Hu, Chuang Gan, and Song Han.
\newblock Efficientvit: Multi-scale linear attention for high-resolution dense prediction, 2023.

\bibitem[Darcet et~al.(2023)Darcet, Oquab, Mairal, and Bojanowski]{darcet2023vision}
Timothée Darcet, Maxime Oquab, Julien Mairal, and Piotr Bojanowski.
\newblock Vision transformers need registers, 2023.

\bibitem[El~Banani et~al.(2024)El~Banani, Raj, Maninis, Kar, Li, Rubinstein, Sun, Guibas, Johnson, and Jampani]{elbanani2024probing}
Mohamed El~Banani, Amit Raj, Kevis-Kokitsi Maninis, Abhishek Kar, Yuanzhen Li, Michael Rubinstein, Deqing Sun, Leonidas Guibas, Justin Johnson, and Varun Jampani.
\newblock {Probing the 3D Awareness of Visual Foundation Models}.
\newblock In \emph{CVPR}, 2024.

\bibitem[Fang et~al.(2023)Fang, Jose, Jain, Schmidt, Toshev, and Shankar]{fang2023data}
Alex Fang, Albin~Madappally Jose, Amit Jain, Ludwig Schmidt, Alexander Toshev, and Vaishaal Shankar.
\newblock Data filtering networks, 2023.

\bibitem[Fang et~al.(2024)Fang, Zhu, Lu, Wang, Molchanov, Cho, Pavone, Han, and Yin]{fang2024vila2vilaaugmentedvila}
Yunhao Fang, Ligeng Zhu, Yao Lu, Yan Wang, Pavlo Molchanov, Jang~Hyun Cho, Marco Pavone, Song Han, and Hongxu Yin.
\newblock $vila^2$: Vila augmented vila, 2024.
\newblock URL \url{https://arxiv.org/abs/2407.17453}.

\bibitem[Fukuda et~al.(2017)Fukuda, Suzuki, Kurata, Thomas, Cui, and Ramabhadran]{Fukuda2017EfficientKD}
Takashi Fukuda, Masayuki Suzuki, Gakuto Kurata, Samuel Thomas, Jia Cui, and Bhuvana Ramabhadran.
\newblock Efficient knowledge distillation from an ensemble of teachers.
\newblock In \emph{Interspeech}, 2017.
\newblock URL \url{https://api.semanticscholar.org/CorpusID:30258763}.

\bibitem[Gadre et~al.(2023)Gadre, Ilharco, Fang, Hayase, Smyrnis, Nguyen, Marten, Wortsman, Ghosh, Zhang, Orgad, Entezari, Daras, Pratt, Ramanujan, Bitton, Marathe, Mussmann, Vencu, Cherti, Krishna, Koh, Saukh, Ratner, Song, Hajishirzi, Farhadi, Beaumont, Oh, Dimakis, Jitsev, Carmon, Shankar, and Schmidt]{gadre2023datacomp}
Samir~Yitzhak Gadre, Gabriel Ilharco, Alex Fang, Jonathan Hayase, Georgios Smyrnis, Thao Nguyen, Ryan Marten, Mitchell Wortsman, Dhruba Ghosh, Jieyu Zhang, Eyal Orgad, Rahim Entezari, Giannis Daras, Sarah Pratt, Vivek Ramanujan, Yonatan Bitton, Kalyani Marathe, Stephen Mussmann, Richard Vencu, Mehdi Cherti, Ranjay Krishna, Pang~Wei Koh, Olga Saukh, Alexander Ratner, Shuran Song, Hannaneh Hajishirzi, Ali Farhadi, Romain Beaumont, Sewoong Oh, Alex Dimakis, Jenia Jitsev, Yair Carmon, Vaishaal Shankar, and Ludwig Schmidt.
\newblock Datacomp: In search of the next generation of multimodal datasets, 2023.

\bibitem[Garrido et~al.(2023)Garrido, Balestriero, Najman, and Lecun]{garrido2023rankme}
Quentin Garrido, Randall Balestriero, Laurent Najman, and Yann Lecun.
\newblock Rankme: Assessing the downstream performance of pretrained self-supervised representations by their rank, 2023.

\bibitem[Girshick(2015)]{girshick2015fast}
Ross Girshick.
\newblock Fast r-cnn.
\newblock In \emph{Proceedings of the IEEE international conference on computer vision}, pp.\  1440--1448, 2015.

\bibitem[Heo et~al.(2019)Heo, Kim, Yun, Park, Kwak, and Choi]{heo2019overhaul}
B.~Heo, J.~Kim, S.~Yun, H.~Park, N.~Kwak, and J.~Choi.
\newblock A comprehensive overhaul of feature distillation.
\newblock In \emph{2019 IEEE/CVF International Conference on Computer Vision (ICCV)}, pp.\  1921--1930, Los Alamitos, CA, USA, nov 2019. IEEE Computer Society.
\newblock \doi{10.1109/ICCV.2019.00201}.
\newblock URL \url{https://doi.ieeecomputersociety.org/10.1109/ICCV.2019.00201}.

\bibitem[Hinton et~al.(2015)Hinton, Vinyals, and Dean]{hinton2015distilling}
Geoffrey Hinton, Oriol Vinyals, and Jeff Dean.
\newblock Distilling the knowledge in a neural network.
\newblock \emph{arXiv preprint arXiv:1503.02531}, 2015.

\bibitem[Hu et~al.(2019)Hu, Dey, Hebert, and Bagnell]{hu2019adaloss}
Hanzhang Hu, Debadeepta Dey, Martial Hebert, and J.~Andrew Bagnell.
\newblock Learning anytime predictions in neural networks via adaptive loss balancing.
\newblock In \emph{Proceedings of the Thirty-Third AAAI Conference on Artificial Intelligence and Thirty-First Innovative Applications of Artificial Intelligence Conference and Ninth AAAI Symposium on Educational Advances in Artificial Intelligence}, AAAI'19/IAAI'19/EAAI'19. AAAI Press, 2019.
\newblock ISBN 978-1-57735-809-1.
\newblock \doi{10.1609/aaai.v33i01.33013812}.
\newblock URL \url{https://doi.org/10.1609/aaai.v33i01.33013812}.

\bibitem[Huang \& Wang(2017)Huang and Wang]{huang2017like}
Zehao Huang and Naiyan Wang.
\newblock Like what you like: Knowledge distill via neuron selectivity transfer.
\newblock \emph{CoRR}, abs/1707.01219, 2017.
\newblock URL \url{http://arxiv.org/abs/1707.01219}.

\bibitem[Ilharco et~al.(2021)Ilharco, Wortsman, Wightman, Gordon, Carlini, Taori, Dave, Shankar, Namkoong, Miller, Hajishirzi, Farhadi, and Schmidt]{ilharco2021openclip}
Gabriel Ilharco, Mitchell Wortsman, Ross Wightman, Cade Gordon, Nicholas Carlini, Rohan Taori, Achal Dave, Vaishaal Shankar, Hongseok Namkoong, John Miller, Hannaneh Hajishirzi, Ali Farhadi, and Ludwig Schmidt.
\newblock Openclip, July 2021.
\newblock URL \url{https://doi.org/10.5281/zenodo.5143773}.

\bibitem[Kanj et~al.(2022)Kanj, Trioux, Coudoux, Gharbi, Corlay, and Kieffer]{kanj2022whitening}
Hind Kanj, Anthony Trioux, François-Xavier Coudoux, Mohamed Gharbi, Patrick Corlay, and Michel Kieffer.
\newblock A comparative study of the whitening methods in linear video coding and transmission schemes.
\newblock In \emph{2022 11th International Symposium on Signal, Image, Video and Communications (ISIVC)}, pp.\  1--6, 2022.
\newblock \doi{10.1109/ISIVC54825.2022.9800738}.

\bibitem[Kessy et~al.(2018)Kessy, Lewin, and Strimmer]{kessy2018whitening}
Agnan Kessy, Alex Lewin, and Korbinian Strimmer.
\newblock {Optimal Whitening and Decorrelation}.
\newblock \emph{The American Statistician}, 72\penalty0 (4):\penalty0 309--314, October 2018.
\newblock \doi{10.1080/00031305.2016.127}.
\newblock URL \url{https://ideas.repec.org/a/taf/amstat/v72y2018i4p309-314.html}.

\bibitem[Kim et~al.(2018)Kim, Park, and Kwak]{kim2018paraphrasing}
Jangho Kim, SeongUk Park, and Nojun Kwak.
\newblock Paraphrasing complex network: Network compression via factor transfer.
\newblock In \emph{Proceedings of the 32nd International Conference on Neural Information Processing Systems}, NIPS'18, pp.\  2765–2774, Red Hook, NY, USA, 2018. Curran Associates Inc.

\bibitem[Kirillov et~al.(2023)Kirillov, Mintun, Ravi, Mao, Rolland, Gustafson, Xiao, Whitehead, Berg, Lo, Dollár, and Girshick]{kirillov2023sam}
Alexander Kirillov, Eric Mintun, Nikhila Ravi, Hanzi Mao, Chloe Rolland, Laura Gustafson, Tete Xiao, Spencer Whitehead, Alexander~C. Berg, Wan-Yen Lo, Piotr Dollár, and Ross Girshick.
\newblock Segment anything, 2023.

\bibitem[Kuhn(1955)]{kuhn1955hungarian}
H.~W. Kuhn.
\newblock The hungarian method for the assignment problem.
\newblock \emph{Naval Research Logistics Quarterly}, 2\penalty0 (1-2):\penalty0 83--97, 1955.
\newblock \doi{https://doi.org/10.1002/nav.3800020109}.
\newblock URL \url{https://onlinelibrary.wiley.com/doi/abs/10.1002/nav.3800020109}.

\bibitem[Lan et~al.(2018)Lan, Zhu, and Gong]{lan2018knowledge}
Xu~Lan, Xiatian Zhu, and Shaogang Gong.
\newblock Knowledge distillation by on-the-fly native ensemble, 2018.

\bibitem[Li et~al.(2024)Li, Zhang, Zhang, Zhang, Li, Li, Ma, and Li]{li2024llavanextinterleavetacklingmultiimagevideo}
Feng Li, Renrui Zhang, Hao Zhang, Yuanhan Zhang, Bo~Li, Wei Li, Zejun Ma, and Chunyuan Li.
\newblock Llava-next-interleave: Tackling multi-image, video, and 3d in large multimodal models, 2024.
\newblock URL \url{https://arxiv.org/abs/2407.07895}.

\bibitem[Liu et~al.(2023)Liu, Li, Li, and Lee]{liu2023improvedllava}
Haotian Liu, Chunyuan Li, Yuheng Li, and Yong~Jae Lee.
\newblock Improved baselines with visual instruction tuning, 2023.

\bibitem[Liu et~al.(2022)Liu, Yang, and Chen]{liu2022normalizedfeaturedistillationsemantic}
Tao Liu, Xi~Yang, and Chenshu Chen.
\newblock Normalized feature distillation for semantic segmentation, 2022.
\newblock URL \url{https://arxiv.org/abs/2207.05256}.

\bibitem[Liu et~al.(2020)Liu, Zhang, and Wang]{liu2020adamultiteachr}
Yuang Liu, Wei Zhang, and Jun Wang.
\newblock Adaptive multi-teacher multi-level knowledge distillation.
\newblock \emph{Neurocomputing}, 415:\penalty0 106--113, nov 2020.
\newblock \doi{10.1016/j.neucom.2020.07.048}.
\newblock URL \url{https://doi.org/10.1016%2Fj.neucom.2020.07.048}.

\bibitem[Mirzadeh et~al.(2019)Mirzadeh, Farajtabar, Li, Levine, Matsukawa, and Ghasemzadeh]{Mirzadeh2019ImprovedKD}
Seyed~Iman Mirzadeh, Mehrdad Farajtabar, Ang Li, Nir Levine, Akihiro Matsukawa, and Hassan Ghasemzadeh.
\newblock Improved knowledge distillation via teacher assistant.
\newblock In \emph{AAAI Conference on Artificial Intelligence}, 2019.
\newblock URL \url{https://api.semanticscholar.org/CorpusID:212908749}.

\bibitem[Oquab et~al.(2023)Oquab, Darcet, Moutakanni, Vo, Szafraniec, Khalidov, Fernandez, Haziza, Massa, El-Nouby, Assran, Ballas, Galuba, Howes, Huang, Li, Misra, Rabbat, Sharma, Synnaeve, Xu, Jegou, Mairal, Labatut, Joulin, and Bojanowski]{oquab2023dinov2}
Maxime Oquab, Timothée Darcet, Théo Moutakanni, Huy Vo, Marc Szafraniec, Vasil Khalidov, Pierre Fernandez, Daniel Haziza, Francisco Massa, Alaaeldin El-Nouby, Mahmoud Assran, Nicolas Ballas, Wojciech Galuba, Russell Howes, Po-Yao Huang, Shang-Wen Li, Ishan Misra, Michael Rabbat, Vasu Sharma, Gabriel Synnaeve, Hu~Xu, Hervé Jegou, Julien Mairal, Patrick Labatut, Armand Joulin, and Piotr Bojanowski.
\newblock Dinov2: Learning robust visual features without supervision, 2023.

\bibitem[Paley(1933)]{Paley1933OnOM}
R~E Paley.
\newblock On orthogonal matrices.
\newblock \emph{Journal of Mathematics and Physics}, 12:\penalty0 311--320, 1933.
\newblock URL \url{https://api.semanticscholar.org/CorpusID:124410493}.

\bibitem[Park \& Kwak(2020)Park and Kwak]{Park2020FeatureLevelEK}
Seonguk Park and Nojun Kwak.
\newblock Feature-level ensemble knowledge distillation for aggregating knowledge from multiple networks.
\newblock In \emph{European Conference on Artificial Intelligence}, 2020.
\newblock URL \url{https://api.semanticscholar.org/CorpusID:220378802}.

\bibitem[Pratt et~al.(1969)Pratt, Kane, and Andrews]{Pratt1969Hadamard}
W.K. Pratt, J.~Kane, and H.C. Andrews.
\newblock Hadamard transform image coding.
\newblock \emph{Proceedings of the IEEE}, 57\penalty0 (1):\penalty0 58--68, 1969.
\newblock \doi{10.1109/PROC.1969.6869}.

\bibitem[Radford et~al.(2021)Radford, Kim, Hallacy, Ramesh, Goh, Agarwal, Sastry, Askell, Mishkin, Clark, Krueger, and Sutskever]{radford2021clip}
Alec Radford, Jong~Wook Kim, Chris Hallacy, Aditya Ramesh, Gabriel Goh, Sandhini Agarwal, Girish Sastry, Amanda Askell, Pamela Mishkin, Jack Clark, Gretchen Krueger, and Ilya Sutskever.
\newblock Learning transferable visual models from natural language supervision.
\newblock In Marina Meila and Tong Zhang (eds.), \emph{Proceedings of the 38th International Conference on Machine Learning}, volume 139 of \emph{Proceedings of Machine Learning Research}, pp.\  8748--8763. PMLR, 18--24 Jul 2021.
\newblock URL \url{https://proceedings.mlr.press/v139/radford21a.html}.

\bibitem[Ranzinger et~al.(2024)Ranzinger, Heinrich, Kautz, and Molchanov]{ranzinger2023amradio}
Mike Ranzinger, Greg Heinrich, Jan Kautz, and Pavlo Molchanov.
\newblock Am-radio: Agglomerative vision foundation model reduce all domains into one.
\newblock In \emph{Proceedings of the IEEE/CVF Conference on Computer Vision and Pattern Recognition (CVPR)}, pp.\  12490--12500, June 2024.

\bibitem[Romero et~al.(2014)Romero, Ballas, Kahou, Chassang, Gatta, and Bengio]{Romero2014FitNetsHF}
Adriana Romero, Nicolas Ballas, Samira~Ebrahimi Kahou, Antoine Chassang, Carlo Gatta, and Yoshua Bengio.
\newblock Fitnets: Hints for thin deep nets.
\newblock \emph{CoRR}, abs/1412.6550, 2014.
\newblock URL \url{https://api.semanticscholar.org/CorpusID:2723173}.

\bibitem[Roy \& Vetterli(2007)Roy and Vetterli]{roy2007effectiverank}
Olivier Roy and Martin Vetterli.
\newblock The effective rank: A measure of effective dimensionality.
\newblock In \emph{2007 15th European Signal Processing Conference}, pp.\  606--610, 2007.

\bibitem[Ruder et~al.(2017)Ruder, Ghaffari, and Breslin]{ruder2017knowledgeadaptationteachingadapt}
Sebastian Ruder, Parsa Ghaffari, and John~G. Breslin.
\newblock Knowledge adaptation: Teaching to adapt, 2017.
\newblock URL \url{https://arxiv.org/abs/1702.02052}.

\bibitem[Sariyildiz et~al.(2024)Sariyildiz, Weinzaepfel, Lucas, Larlus, and Kalantidis]{sariyildiz2024unicuniversalclassificationmodels}
Mert~Bulent Sariyildiz, Philippe Weinzaepfel, Thomas Lucas, Diane Larlus, and Yannis Kalantidis.
\newblock Unic: Universal classification models via multi-teacher distillation, 2024.
\newblock URL \url{https://arxiv.org/abs/2408.05088}.

\bibitem[Shang et~al.(2024)Shang, Schmeckpeper, May, Minniti, Kelestemur, Watkins, and Herlant]{shang2024theia}
Jinghuan Shang, Karl Schmeckpeper, Brandon~B. May, Maria~Vittoria Minniti, Tarik Kelestemur, David Watkins, and Laura Herlant.
\newblock Theia: Distilling diverse vision foundation models for robot learning.
\newblock In \emph{8th Annual Conference on Robot Learning}, 2024.
\newblock URL \url{https://openreview.net/forum?id=ylZHvlwUcI}.

\bibitem[Sun et~al.(2021)Sun, Panda, Chen, Oliva, Feris, and Saenko]{sun2021dynamic}
X.~Sun, R.~Panda, C.~Chen, A.~Oliva, R.~Feris, and K.~Saenko.
\newblock Dynamic network quantization for efficient video inference.
\newblock In \emph{2021 IEEE/CVF International Conference on Computer Vision (ICCV)}, pp.\  7355--7365, Los Alamitos, CA, USA, oct 2021. IEEE Computer Society.
\newblock \doi{10.1109/ICCV48922.2021.00728}.
\newblock URL \url{https://doi.ieeecomputersociety.org/10.1109/ICCV48922.2021.00728}.

\bibitem[Sylvester(1867)]{Sylvester1867LXTO}
James Sylvester.
\newblock Lx. thoughts on inverse orthogonal matrices, simultaneous signsuccessions, and tessellated pavements in two or more colours, with applications to newton's rule, ornamental tile-work, and the theory of numbers.
\newblock \emph{Philosophical Magazine Series 1}, 34:\penalty0 461--475, 1867.
\newblock URL \url{https://api.semanticscholar.org/CorpusID:118420043}.

\bibitem[Wei et~al.(2022{\natexlab{a}})Wei, Hu, Xie, Zhang, Cao, Bao, Chen, and Guo]{wei2022contrastive}
Yixuan Wei, Han Hu, Zhenda Xie, Zheng Zhang, Yue Cao, Jianmin Bao, Dong Chen, and Baining Guo.
\newblock Contrastive learning rivals masked image modeling in fine-tuning via feature distillation, 2022{\natexlab{a}}.

\bibitem[Wei et~al.(2022{\natexlab{b}})Wei, Hu, Xie, Zhang, Cao, Bao, Chen, and Guo]{wei2022featuredistillation}
Yixuan Wei, Han Hu, Zhenda Xie, Zheng Zhang, Yue Cao, Jianmin Bao, Dong Chen, and Baining Guo.
\newblock Contrastive learning rivals masked image modeling in fine-tuning via feature distillation, 2022{\natexlab{b}}.

\bibitem[Wightman(2019)]{rw2019timm}
Ross Wightman.
\newblock Pytorch image models.
\newblock \url{https://github.com/rwightman/pytorch-image-models}, 2019.

\bibitem[Yang et~al.(2020)Yang, Shou, Gong, Lin, and Jiang]{yang2020modelcompression}
Ze~Yang, Linjun Shou, Ming Gong, Wutao Lin, and Daxin Jiang.
\newblock Model compression with two-stage multi-teacher knowledge distillation for web question answering system.
\newblock In \emph{Proceedings of the 13th International Conference on Web Search and Data Mining}, WSDM '20, pp.\  690–698, New York, NY, USA, 2020. Association for Computing Machinery.
\newblock ISBN 9781450368223.
\newblock \doi{10.1145/3336191.3371792}.
\newblock URL \url{https://doi.org/10.1145/3336191.3371792}.

\bibitem[You et~al.(2017)You, Xu, Xu, and Tao]{you2017multiteacher}
Shan You, Chang Xu, Chao Xu, and Dacheng Tao.
\newblock Learning from multiple teacher networks.
\newblock In \emph{Proceedings of the 23rd ACM SIGKDD International Conference on Knowledge Discovery and Data Mining}, KDD '17, pp.\  1285–1294, New York, NY, USA, 2017. Association for Computing Machinery.
\newblock ISBN 9781450348874.
\newblock \doi{10.1145/3097983.3098135}.
\newblock URL \url{https://doi.org/10.1145/3097983.3098135}.

\bibitem[Yuan et~al.(2020)Yuan, Shou, Pei, Lin, Gong, Fu, and Jiang]{yuan2020reinforced}
Fei Yuan, Linjun Shou, Jian Pei, Wutao Lin, Ming Gong, Yan Fu, and Daxin Jiang.
\newblock Reinforced multi-teacher selection for knowledge distillation, 2020.

\bibitem[Zagoruyko \& Komodakis(2017)Zagoruyko and Komodakis]{zagoruyko2017paying}
Sergey Zagoruyko and Nikos Komodakis.
\newblock Paying more attention to attention: Improving the performance of convolutional neural networks via attention transfer.
\newblock In \emph{5th International Conference on Learning Representations, {ICLR} 2017, Toulon, France, April 24-26, 2017, Conference Track Proceedings}. OpenReview.net, 2017.
\newblock URL \url{https://openreview.net/forum?id=Sks9\_ajex}.

\bibitem[Zbontar et~al.(2021)Zbontar, Jing, Misra, LeCun, and Deny]{zbontar2021barlow}
Jure Zbontar, Li~Jing, Ishan Misra, Yann LeCun, and St{\'e}phane Deny.
\newblock Barlow twins: Self-supervised learning via redundancy reduction.
\newblock \emph{arXiv preprint arXiv:2103.03230}, 2021.

\bibitem[Zhai et~al.(2023{\natexlab{a}})Zhai, Likhomanenko, Littwin, Busbridge, Ramapuram, Zhang, Gu, and Susskind]{zhai2023stabilizing}
Shuangfei Zhai, Tatiana Likhomanenko, Etai Littwin, Dan Busbridge, Jason Ramapuram, Yizhe Zhang, Jiatao Gu, and Joshua~M Susskind.
\newblock Stabilizing transformer training by preventing attention entropy collapse.
\newblock In \emph{International Conference on Machine Learning}, pp.\  40770--40803. PMLR, 2023{\natexlab{a}}.

\bibitem[Zhai et~al.(2023{\natexlab{b}})Zhai, Mustafa, Kolesnikov, and Beyer]{zhai2023sigmoid}
Xiaohua Zhai, Basil Mustafa, Alexander Kolesnikov, and Lucas Beyer.
\newblock Sigmoid loss for language image pre-training.
\newblock \emph{arXiv preprint arXiv:2303.15343}, 2023{\natexlab{b}}.

\bibitem[Zhao et~al.(2022)Zhao, Sun, Dong, Chen, and Dong]{zhao2022collabteaching}
Haoran Zhao, Xin Sun, Junyu Dong, Changrui Chen, and Zihe Dong.
\newblock Highlight every step: Knowledge distillation via collaborative teaching.
\newblock \emph{IEEE Transactions on Cybernetics}, 52\penalty0 (4):\penalty0 2070--2081, 2022.
\newblock \doi{10.1109/TCYB.2020.3007506}.

\bibitem[Zhou et~al.(2022)Zhou, Wei, Wang, Shen, Xie, Yuille, and Kong]{zhou2022ibot}
Jinghao Zhou, Chen Wei, Huiyu Wang, Wei Shen, Cihang Xie, Alan Yuille, and Tao Kong.
\newblock Image {BERT} pre-training with online tokenizer.
\newblock In \emph{International Conference on Learning Representations}, 2022.
\newblock URL \url{https://openreview.net/forum?id=ydopy-e6Dg}.

\bibitem[Zhu et~al.(2024)Zhu, Zhou, Wang, Cao, Han, Hou, and Xu]{zhu2024unit}
Yi~Zhu, Yanpeng Zhou, Chunwei Wang, Yang Cao, Jianhua Han, Lu~Hou, and Hang Xu.
\newblock Unit: Unifying image and text recognition in one vision encoder, 2024.
\newblock URL \url{https://arxiv.org/abs/2409.04095}.

\bibitem[Zuchniak(2023)]{zuchniak2023multiteacher}
Konrad Zuchniak.
\newblock Multi-teacher knowledge distillation as an effective method for compressing ensembles of neural networks, 2023.

\end{thebibliography}
\bibliographystyle{preprint}

\clearpage
\setcounter{page}{1}

\onecolumn

\appendix

\section{Appendix}

\subsection{Hadamard Matrices}

\subsubsection{Constructing Hadamard Matrices}\label{sec:general_hadamard_matrices}

Sylvester's construction gives us a convenient way to construct a Hadamard matrix when $C$ is a power of 2. Unfortunately, many of the $C$s we care about aren't such a power. More generally, the Hadamard Conjecture hypothesizes that there exists a valid Hadamard matrix for any $C$ that is divisible by 4. If true, then there are significantly more valid matrices, and in particular, common deep learning choices will be a multiple of 4. While not proven in general, the literature has found a way to construct many non-power-of-2 sized matrices using some of the following rules:

\begin{itemize}
    \item If $\mathbf{H}_n$ and $\mathbf{H}_m$ are Hadamard matrices, then $\mathbf{H}_n \bigotimes \mathbf{H}_m$ is also a Hadamard matrix.
    \item If $3 \equiv q^k \mod 4$ for some prime $q$ and integer $k > 0$, then we can use Paley's first construction \cite{Paley1933OnOM} to produce a Hadamard matrix of size $q + 1$.
    \item If $1 \equiv q^k \mod 4$ for some prime $q$ and integer $k > 0$, then we can use Paley's second construction to produce a Hadamard matrix of size $2\left(q + 1\right)$.
\end{itemize}

where $\bigotimes$ is the Kronecker product. For our purposes, there are common feature dimensions that we want to be able to produce:
\begin{itemize}
    \item ViT-B: 768 $\left[\mathbf{S}(2) \bigotimes \mathbf{P}_1(384))\right]$
    \item ViT-L: 1024 $\left[\mathbf{S}(1024)\right]$
    \item SigLIP-L: 1152 $\left[\mathbf{S}(32) \bigotimes \mathbf{P}_2(36)\right]$
    \item ViT-H: 1280 $\left[\mathbf{S}(64) \bigotimes \mathbf{P}_1(20)\right]$
    \item ViT-g: 1408 $\left[\mathbf{S}(32) \bigotimes \mathbf{P}_1(44)\right]$
\end{itemize}

Where $\mathbf{P}_i(x)$ is a Paley construction $i$ of size $x$, and $\mathbf{S}(x)$ is a Sylvester construction of size $x$. In the case of Sylvester, we're referring to when $2^k = x$ for some $k \in \mathbb{N}_{0}$. For $\mathbf{P}_1(384)$, we have the prime $q=383$, which $3 \equiv 383^1 \mod 4$. For $1280$, we can use (possibly among other options) $\mathbf{P}_1(1280)$ as we have $q=1279$, and thus $3 \equiv 1279^1 \mod 4$, or the compound version shown above. Finally, for $P(44)$ we have $q=43$ and $3 \equiv 43^1 \mod 4$. So, by some stroke of luck, we have known constructions of Hadamard matrices for the major ViT widths. There are even more methods for constructing these matrices, and at the time of this writing, the smallest unknown Hadamard matrix is 668. While not exhaustive, for our purposes, the Sylvester and Paley constructions were sufficient to cover the models we studied.

\subsubsection{Using Hadamard Matrices for Noise Suppression / Quantization}

While unrelated to our work of using Hadamard matrices to perform statistical normalization, the recently proposed QuaRot~\cite{ashkboos2024quarotoutlierfree4bitinference} finds a different application of this structured matrix to eliminate activation outliers, making low-bit quantization much more effective.

\subsection{Hadamard Whitening}

\subsubsection{Proof of HCA uniform error profile}\label{sec:hca_error_profile}

Referring to \eqref{eq:error_hadamard}:

\begin{equation}
    \bm{\epsilon}_{\text{hada}} = \mathbf{U\Lambda}^{\frac{1}{2}}\mathbf{H}^\intercal \epsilon
    \tag{\ref{eq:error_hadamard} revisited}
    \label{eq:error_hadamard_rev}
\end{equation}

we demonstrate that each column of $\mathbf{U\Lambda}^\frac{1}{2}\mathbf{H}^\intercal$ has identical magnitude, and further, that an error step of size $\delta$ along any single dimension has identical magnitude in the original space.

\begin{equation}
    \mathbf{\Lambda}^\frac{1}{2}\mathbf{H}^\intercal = \frac{1}{\sqrt{C}} \begin{bmatrix}
        \pm \sqrt{\lambda_1} & \ldots & \pm \sqrt{\lambda_1} \\
        \pm \sqrt{\lambda_2} & \ldots & \pm \sqrt{\lambda_2} \\
        \vdots               & \ddots & \vdots \\
        \pm \sqrt{\lambda_C} & \ldots & \pm \sqrt{\lambda_C}
    \end{bmatrix}
\end{equation}

\begin{equation}
    \left\lVert\mathbf{\Lambda}^\frac{1}{2}\mathbf{H}^\intercal\right\rVert_{\left[:,j\right]} = \sqrt{\sum_{c=1}^C \frac{\lambda_c}{C}} \quad \forall j \in C \label{eq:hada_col_mag}
\end{equation}

where $\left\lVert \cdot \right\rVert_{\left[:,j\right]}$ denotes the norm of column $j$. Equation \ref{eq:hada_col_mag} shows that each column vector has an identical magnitude. Because orthogonal transforms are magnitude preserving, we also get

\begin{equation}
    \left\lVert\mathbf{U\Lambda}^\frac{1}{2}\mathbf{H}^\intercal\right\rVert_{\left[:,j\right]} = \sqrt{\sum_{c=1}^C \frac{\lambda_c}{C}} \quad \forall j \in C
\end{equation}

In particular, this means that for some $\mathbf{\Delta}_r \in \delta\left[\pm \mathbbm{1}_{[r=1]},\ldots, \pm \mathbbm{1}_{[r=C]}\right]^\intercal$ with $\mathbbm{1}_{r=x}$ representing the Kronecker delta for whether $r=x$ and $\delta \in \mathbb{R}_{+}$ (e.g. $\mathbf{\Delta}_r$ is a one-hot column vector with a $\pm 1$ at position $r$ multiplied by some positive real $\delta$), then

\begin{equation}
    \left\lVert \left(\mathbf{U\Lambda}^\frac{1}{2}\mathbf{H}^\intercal\right) \mathbf{\Delta}_r \right\rVert = \delta\sqrt{\sum_{c=1}^C \frac{\lambda_c}{C}}  \quad \forall r \in C
    \label{eq:hadamard_delta_error}
\end{equation}

In words, an error step of size $\delta$ along any single axis $r$ in whitened space will be scaled by

\begin{equation}
    \sqrt{\frac{1}{C}\sum_{c=1}^C \lambda_c}
    \label{eq:hadamard_error_mag}
\end{equation}

when projecting back into the original space. So each dimension being learned by the student has the same magnitude of effect in the original teacher space. Our hypothesis is that this should improve learning dynamics as there is no implicitly more important dimension to match than any other, compared with PCA-W which places the most importance on the first dimension, and so on. Note that an arbitrary error vector of magnitude $\delta$ does not have this property since $\mathbf{U\Lambda}^\frac{1}{2}\mathbf{H}^\intercal$ is not orthogonal in general.

Incidentally, \eqref{eq:hadamard_error_mag} is identical to \eqref{eq:ship_phi} which is the radius of the denormalized unit error circle for PHI-S. This means that at any $\bm{\Delta}_r$ with $\delta = 1$, the error magnitude is identical between the two normalization methods. We visualize this when $C=2$ in figure \ref{fig:radial_error_magnitudes} by looking at where the blue and purple curves intersect.

\subsection{Teacher Effective Ranks}

We apply the RankMe~\cite{garrido2023rankme} algorithm to a handful of models, including the set of teachers used for training. While it was technically only designed for SSL models (like DINOv2), it may still lend insight into why whitening didn't empirically work well. The results are in table \ref{tab:teacher_effective_ranks}, where we show that the effective ranks for all teachers are much smaller than their number of channels. It is also interesting to consider whether agglomerative models work because the teachers aren't effectively full rank, suggesting that we can pack more information from other teachers into a student of equivalent or larger size than the teachers. More investigation is needed to understand why the RADIO models (both AM-RADIO and ours) seem to be lower rank than their counterparts of equivalent size.

\begin{table*}[]
    \centering
    \resizebox{0.4\linewidth}{!}{
    \begin{tabular}{r|ccc}
        \textbf{Model} & C     & \textbf{RankMe} \\
        \hline
        DINOv2-b-reg   & 768   & 685.52  \\
        PHI-S-RADIO-B   & 768   & 645.38  \\
        \hline
        DINOv2-l-reg   & 1024  & 906.48  \\
        PHI-S-RADIO-L   & 1024  & 859.23  \\
        \hline
        SigLIP   (-L)  & 1152  & 910.97  \\
        \hline
        DFN CLIP (-H)  & 1280  & 1081.69 \\
        SAM (-H)       & 1280  & 776.28\\
        AM-RADIO (-H)  & 1280  & 1043.84 \\
        \hline
        DINOv2-g-reg   & 1536  & 1342.55
    \end{tabular}
    }
    \caption{The effective rank estimates for the spatial features of various models using the RankMe \cite{garrido2023rankme} algorithm. As can be seen, the effective rank is much smaller than $C$, meaning that whitening methods will have a large number of dimensions with very small variance. This likely helps to explain why the whitening methods produced the students with the highest losses. }
    \label{tab:teacher_effective_ranks}
\end{table*}

\subsection{Teacher Distribution Statistics}

In table \ref{tab:teacher_activation_stats} we show statistics about the distributions of the teachers. We can see that they are not mean centered, and also that their standard deviations are very different, both globally, and per-channel.

\begin{table*}[!ht]
    \centering
    \resizebox{0.65\linewidth}{!}{
    \begin{tabular}{lcccccc}
        \toprule
        Model & \multicolumn{4}{c}{Per Channel} & \multicolumn{2}{c}{Global} \\
        \cmidrule(lr){2-5} \cmidrule(lr){6-7}
         & \multicolumn{2}{c}{Mean} & \multicolumn{2}{c}{Std} & Mean & Std \\
        \cmidrule(lr){2-3} \cmidrule(lr){4-5}
         & Min & Max & Min & Max & & \\
        \midrule
        DFN CLIP & -0.1689  & 0.1385  & 0.0105 & 0.1334  & 0.0049 & 0.0286 \\
        SigLIP   & -6.8789  & 31.25   & 0.3813 & 21.6875 & 0.0211 & 1.8389 \\
        DINOv2   & -3.3945  & 4.293   & 0.3918 & 4.3008  & 0.0055 & 1.3496 \\
        SAM      & -62.0312 & 19.1719 & 2.6953 & 31.6094 & 1.1475 & 5.4688 \\
        \bottomrule
    \end{tabular}
    }
    \caption{Activation statistics for various teachers. Here we can see that each of the teachers' distributions have very different standard deviations (Global). We can also see that different channels for a given teacher have very different means and standard deviations (Per Channel). Taking SAM as an example: The smallest mean value channel has value $-62.0312$, and largest channel $19.1719$. Similarly, the channel with smallest standard deviation has $2.6953$, and channel with largest has $31.6094$.}
    \label{tab:teacher_activation_stats}
\end{table*}

\subsection{Additional PHI-S Insights}

\subsubsection{Role of PCA}\label{apdx:ship_role_of_pca}

Because rotations are magnitude preserving (and thus variance preserving), with $\cov{\mathbf{Y}} = \mathbf{U\Lambda U}^\intercal$, then $\trace\left(\cov{\mathbf{Y}}\right) = \trace\left(\mathbf{\Lambda}\right) = \sum_i^C \lambda_i$. This means that the normalization ($\alpha$) derived in \eqref{eq:ship_proof} is a constant with respect to the distribution, invariant to any orthogonal transform that's applied to it. It will always be $\sqrt{\frac{1}{C} \sum_i^C \lambda_i}$. And so, we have that

\begin{align}
    \cov{\mathbf{HY}} &= \mathbf{H}\left(\mathbf{U\Lambda U}^\intercal\right) \mathbf{H}^\intercal
\end{align}

where the $\mathbf{U}$ is preventing the $\mathbf{H}$ from evenly distributing the variance in $\bm{\Lambda}$, unless $\mathbf{U} = \mathbf{I}$, or worst case $\mathbf{U} = \mathbf{H}^\intercal$ in which case applying $\mathbf{H}$ would result in $\bm{\Lambda}$ variance, the opposite of what we want. So, we don't need PCA to find the scale $\alpha$ to normalize the distribution, but we do need it to find the orthogonal transform $\mathbf{HU}^\intercal$ which results in a dimensionally balanced  distribution.

\begin{align}
    \cov{\mathbf{HU}^\intercal \mathbf{Y}} &= \left(\mathbf{HU}^\intercal\right)\left(\mathbf{U\Lambda U}^\intercal\right) \left(\mathbf{UH}^\intercal\right) \\
                      &= \mathbf{H}\bm{\Lambda}\mathbf{H}^\intercal
\end{align}

\subsubsection{Comparison between Global Standardization and PHI-S}\label{sec:gstd_vs_phis}

In table \ref{tab:iso_std_scales} we show what the normalization scalars are for each teacher distribution. Because both methods use a single value to rescale the distribution, it's useful to see how they treat the same distribution. Notably, PHI-S uses a larger scale for all of the teacher distributions. It's also worth noting that the difference in scales is not constant across the distributions. Both methods are invariant to the orientation of the distribution, thus these scalars are unique properties of the distribution.

\begin{table}[!h]
    \centering
    \begin{tabular}{r|cc}
        \textbf{Model} & $\alpha_{\text{gs}}$ & $\alpha_\text{ship}$  \\
        \hline
        DFN CLIP & 35.02  & 41.41  \\
        SigLIP   &  0.53  &  0.65  \\
        DINOv2   &  0.73  &  0.76  \\
        SAM      &  0.19  &  0.21  \\
    \end{tabular}
    \caption{Comparison of scales between global standardization \eqref{eq:global_standardize} and PHI-S standardization \eqref{eq:ship}. We get $\alpha_{\text{gs}} = \frac{1}{\sigma_g}$, which is the scaling factor for global standardization.}
    \label{tab:iso_std_scales}
\end{table}

A natural question arises: Why is $\alpha_{\text{gs}} \neq \alpha_{\text{ship}}$?

Recall that
\begin{align}
    \alpha_{\text{ship}} = \phi^{-1} &= \left(\frac{1}{C} \sum_i^C \lambda_i\right)^{-\frac{1}{2}}
    \tag{\ref{eq:ship_phi} \& \ref{eq:ship_proof} revisited} \\
                         &= \left(\frac{1}{C} \trace\left(\cov{\mathbf{Y}}\right)\right)^{-\frac{1}{2}}
\end{align}

And also how in section \ref{apdx:ship_role_of_pca} we showed that $\alpha_{\text{ship}}$ is invariant to any orthogonal transform on the distribution. For global standardization, we reinterpret the multivariate distribution as univariate, thus we get scalar $\mu_g$ and $\sigma_g$, global mean and global standard deviation respectively. For the multivariate distribution, we have $\bm{\mu}$, the vector of means for each dimension. We can equivalently write the computation of $\sigma_g$ as

\begin{equation}
    \sigma_g = \sqrt{\frac{1}{NC-1} \sum_i^N \sum_c^C \left(y_{i,c} - \mu_g \right)^2 }
\end{equation}

and then for $\phi_{\text{ship}}$ we have

\begin{equation}
    \phi_{\text{ship}} = \sqrt{\frac{1}{N(C-1)} \sum_i^N \sum_c^C \left(y_{i,c} - \mu_c \right)^2  }
\end{equation}

therefore, when $\mu_c = \mu_g \quad \forall c \in C$, then $\lim_{N\to\infty} \sigma_g = \lim_{N\to\infty}\phi_{\text{ship}}$. Meaning that, as long as the mean for each dimension of the distribution is the same, then Global Standardization and PHI-S will arrive at nearly the same scaling constant when N is large, and thus only differ by rotation. A trivial example is when the distribution is already mean centered on every dimension. We show in table \ref{tab:teacher_activation_stats} that none of the teachers we studied have uniform mean per channel, which is why the methods end up with different scaling constants.

\subsubsection{How similar are the original teacher distributions to the PHI-S distribution?}\label{sec:orig_ship_dist_similarity}

The main property of PHI-S is that it rotates the distribution in such a way that the standard deviation for each channel is identical, allowing us to standardize the distribution in this rotated space using a single scalar. From \eqref{eq:ship_proof}, if the two distributions are aligned, then $\mathbf{HU}^\intercal = \mathbf{I^*}$ with $\mathbf{I^*}$ being some permutation of $\mathbf{I}$. We measure the deviation from this ideal by computing $\text{abs}\left(\mathbf{HU}^\intercal\right)$, and then using the Hungarian algorithm \cite{kuhn1955hungarian} to find the best match of basis vectors $\mathbf{U}_\text{align}^\intercal$, and finally calculating statistics on $\text{diag} \circ \text{abs}\left(\mathbf{HU}_\text{align}^\intercal\right)$, which we show in table \ref{tab:eig_ship_similarity}. We observe that in general, the original distribution is quite unlike that of PHI-S, where at most one basis vector is mostly aligned, but otherwise $\mathbf{H}$ and $\mathbf{U}$ are highly dissimilar.

\begin{table}[]
\centering
\begin{tabular}{r|cccc}
    \textbf{Model} & \textbf{Mean} & \textbf{Min} & \textbf{Max} & $\mathbf{\# > 0.75}$ \\ 
    \hline
    DFN CLIP       & 0.0229 & 0.0000 & 0.9916 & 1 \\
    SigLIP         & 0.0246 & 0.0001 & 0.7864 & 1 \\
    DINOv2         & 0.0203 & 0.0000 & 0.9807 & 1 \\
    SAM            & 0.0226 & 0.0000 & 0.1128 & 0 \\
\end{tabular}
\caption{Measuring how aligned the original teacher distribution is with the PHI-S distribution. Refer to section \ref{sec:orig_ship_dist_similarity} for how this is calculated.}
\label{tab:eig_ship_similarity}
\end{table}

\subsubsection{Degenerate Rank Distributions}

There are additional useful properties for the PHI-S transform, particularly when the original data distribution is not full rank, which is almost certainly the case with deep learning models (table \ref{tab:teacher_effective_ranks}, \cite{garrido2023rankme}). Namely, with the whitening procedures, they will create extreme scale distortions on the zero or negligible eigenvalue dimensions, which can cause the student model to waste too many resources optimizing negligible dimensions. Vanilla standardization also suffers from the same effect, but it may be less aggressive as it's not using PCA which disentangles the dimensions, rather its sensitivity to this problem relies on the orientation of the original distribution. PHI-S, on the other hand, will be well behaved whenever $\text{Rank}\left(\mathbf{Y}\right) \geq 1$ because the rotation will place the same amount of variance on every output dimension. We use the definition in \cite{roy2007effectiverank} for Rank, the effective rank.

Every normalization method, except for PHI-S and Global Standardization, is vulnerable to when $\text{Rank}\left(\mathbf{Y}\right) \leq C$, which we illustrate in the 2-dimensional case:

Let $\mathbf{X} \in \mathbb{R}^{2 \times N}$ be a data distribution with covariance $\cov{\mathbf{X}} = \begin{bmatrix}
    1 & 0 \\ 0 & \epsilon
\end{bmatrix}$. Because standardization \ref{sec:standardization} requires division by the variance in each dimension, then $\lim_{\epsilon\to 0} \sigma_y^{-1} = \infty$. For the whitening methods \ref{sec:whitening}, the diagonalization of $\cov{\mathbf{X}}$ produces $\mathbf{\Lambda} = \diagembed(1, \epsilon)$. The whitening methods then require $\mathbf{\Lambda}^{-\frac{1}{2}}$ which again produces a division by 0 for the y-dimension. Because the PHI-S method operates on the mean eigenvalue, it will have $\lim_{\epsilon\to 0}\alpha = \frac{1}{\sqrt{0.5}} = \sqrt{2}$, which is well defined. While this is a trivial example, the implications are meaningful on real data too, which we show in table \ref{tab:teacher_effective_ranks}.

\subsection{Normalization without runtime penalty}\label{sec:denormalize}

All of the normalization methods introduce extra computation in the form of mean subtraction and some scaling method. Because the teacher adaptors for our model all end with a $y=\mathbf{W'x}+\mathbf{b}'$ linear layer, we can modify this layer after training to produce outputs in the original teacher space. Let $\mathbf{y}'$ be the normalized teacher outputs, $\mathbf{y}$ be the original teacher outputs, $\mathbf{x'}^{(n)}$ be the output of the student matching the normalized teacher (at layer $n$), and we seek to produce a $\mathbf{x}^{(n)}$ that approximates the original teacher distribution. With this, we have:

\begin{align}
    \mathbf{x'}^{(n)} &= \mathbf{W'}\mathbf{x'}^{(n-1)} + \mathbf{b'} \\
    \mathbf{x}^{(n)} &= \mathbf{\Theta}\left(\mathbf{W'x}^{(n-1)} + \mathbf{b'}\right) + \bm{\mu} \\
                     &= \mathbf{\Theta}\mathbf{W'}\mathbf{x}^{(n-1)} + \mathbf{\Theta b'} + \bm{\mu} \\
    \mathbf{W} &= \mathbf{\Theta W'} \\
    \mathbf{b} &= \mathbf{\Theta b'} + \bm{\mu} \\
    \mathbf{x}^{(n)} &= \mathbf{W} \mathbf{x'}^{(n-1)} + \mathbf{b}
    \label{eq:standard_correct}
\end{align}

with $\mathbf{W'}$ and $\mathbf{b'}$ being the weights and bias of the final linear layer of the model respectively. $\mathbf{\Theta}$ and $\mathbf{\mu}$ are the linear correction parameters for the given normalization method.

\begin{itemize}
    \item \textbf{Global Standardize} (\ref{sec:global_standardize}): \begin{equation}
        \mathbf{\Theta} = \mathbf{I}\sigma_g, \quad \bm{\mu} = \bm{1}\mu_g
    \end{equation}
    \item \textbf{Standardize} (\ref{sec:standardize}): \begin{equation}
        \mathbf{\Theta} = \diagembed\left(\sigma_1, ..., \sigma_C\right)
    \end{equation}
    \item \textbf{PCA Whitening} (\ref{sec:zca_whiten}): \begin{equation}
        \mathbf{\Theta} = \mathbf{U\Lambda}^{\frac{1}{2}}
    \end{equation}
    \item \textbf{ZCA Whitening} (\ref{sec:zca_whiten}): \begin{equation}
        \mathbf{\Theta} = \cov{\mathbf{Y}}^{\frac{1}{2}}
    \end{equation}
    \item \textbf{Hadamard Whitening} (\ref{sec:hadamard_whiten}): \begin{equation}
        \mathbf{\Theta} = \mathbf{U\Lambda}^{\frac{1}{2}}\mathbf{H}^\intercal
    \end{equation}
    \item \textbf{PHI-S} (\ref{sec:ship}): \begin{equation}
        \mathbf{\Theta} = \phi\mathbf{UH}^\intercal
    \end{equation}
\end{itemize}

\subsection{Implementation Details}\label{apdx:implementation_details}

In addition to all of the ablations, we also reported PHI-S-RADIO-B and PHI-S-RADIO-L models (Table \ref{tab:headline_metrics}). To produce these models, we add 2 more training stages on top of that in section \ref{sec:ablation_impl_details} as follows:

\begin{itemize}
    \item Stage 1 - Outlined in section \ref{sec:ablation_impl_details} (32 A100 GPUs for 40 hours)
    \item Stage 2 - Increase the student resolution to 432 and train for 300k more steps (64 A100 GPUs for 64 hours)
    \item Stage 3 - Add a ``high res'' set of partitions. Similar to AM-RADIO, we set the batch size to 128 for hi-res while keeping 1024 for low-res. We again train for another 300k steps. (128 A100 GPUs for 68 hours)
\end{itemize}

The multi-stage strategy results in 14,080 total GPU hours for the ViT-B/16 model. If we were to instead train stage 3 for 600k steps (AM-RADIO recipe), it would result in 17,408 total GPU hours. Hyperparameters are shown in table \ref{tab:hparams}.

We employ spectral reparametrization \cite{zhai2023stabilizing} for all stages of training. We've found this to be particularly helpful for stage 3 training when dealing with high resolution. In order to encourage the spectral norm to be small, we ensure that weight decay is applied to the rescaling parameter.

\begin{table}[]
    \centering
    \begin{tabular}{c|ccc}
        \bf{Hyperparameter} & \bf{Stage 1} & \bf{Stage 2} & \bf{Stage 3} \\
        \hline
        Dataset             & DC1B         & DC1B         & DC1B         \\
        Batch Size          & 1024         & 1024         & 1152         \\
        GPUs                & 32           & 64           & 128          \\
        Steps               & 300,000      & 300,000      & 300,000      \\
        LR                  & 1e-3         & 1e-3         & 1e-3         \\
        LR Schedule         & cosine       & cosine       & cosine       \\
        Weight Decay        & 0.02         & 0.02         & 0.02         \\
        Dist Est. Steps     & 3,000        & 3,000        & 3,000        \\
        Frozen Body Steps   & 5,000        & 5,000        & 5,000        \\
        Optimizer           & LAMB         & LAMB         & LAMB
    \end{tabular}
    \caption{
        Hyperparameter table for the training stages. For each stage, we ``restart'' the learning rate schedule at $1e-3$. ``Dist Est. Steps'' describes the number of steps we use at the beginning of the training stage to estimate the teacher data distributions. We reset these estimates for each stage, as the change in resolution may impact these distributions. We also freeze the trunk of the model for ``Frozen Body Steps'' at the start of each stage to allow for the heads to adjust to the new distributions, and also because these distributions may drastically change early on as the estimates are refined. Particularly, methods that rely on matrix diagonalization can undergo major shifts as PyTorch's implementation of torch.eigh() is not particularly stable under small changes to the covariance matrix. ZCA whitening \textit{is} stable upon small estimate updates, owing to the fact that the $\mathbf{U}^\intercal$ rotation is inverted after rescaling, so any permutation of eigenvectors is also negated. DC1B stands for ``DataComp-1B'' \cite{gadre2023datacomp}, from which we only use the images.
    }
    \label{tab:hparams}
\end{table}

\subsection{Raw Metrics}

\subsubsection{Adaptive Balancing}

In AM-RADIO, the authors also explore the use of AdaLoss \cite{hu2019adaloss}, which sets each loss term to be approximately $1$ by dividing the term by the exponential moving average of itself. We explore using this balancing mechanism, both as a standalone (e.g. Baseline + AdaLoss), as well as in conjunction with PHI-S. Table \ref{tab:adaloss_raw_mse} shows the teacher MSEs, and table \ref{tab:vitb_method_ranks_adaloss} shows the benchmark ranks with AdaLoss included. In general, AdaLoss places much more weight on the summary losses, resulting in outsized gains in classification tasks at the expense of dense tasks. We also find that AdaLoss+PHI-S is better than AdaLoss alone.

\begin{table}[]
    \centering
    \resizebox{0.65\linewidth}{!}{
    \begin{tabular}{r|cccc}
        \B{Method} $\downarrow$   & \B{DFN CLIP} ($\cdot 1^{-4}$)    & \B{SigLIP} & \B{DINOv2} & \B{SAM}    \\
        \hline                                                                    
        Ada - MSE     & 4.7790           & 1.9260      & 0.9591      & 8.7500      \\
        Ada - PHI-S    & 4.7750           & 1.9260      & 0.9585      & 8.6960
    \end{tabular}
    }
    \caption{Mean Squared Error for matching the teachers with a ViT-B/16 student using AdaLoss, either normally (Ada - MSE), or in conjunction with PHI-S.}
    \label{tab:adaloss_raw_mse}
\end{table}

\begin{table*}[h]
    \centering
    \resizebox{\linewidth}{!}{
    \begin{tabular}{r|cccccc|ccc}
                         \multirow{2}{*}{\B{Method}} & \B{Teacher}     & \B{Classif-}  & \B{Segment-}& \B{SAM}  & \B{LLaVA}& \B{Probe}& \multirow{2}{*}{\B{Avg}}  & \B{Avg}     & \B{Avg No}   \\
                          & \B{MSE}         & \B{ication}   & \B{ation}   & \B{COCO} & \B{1.5}  & \B{3D}   &                           & \B{No COCO} & \B{MSE/COCO} \\
                     \hline
                   & \multicolumn{9}{c}{\B{Baselines}}                                                                                                            \\
                     \hline
        MSE           & 7.75            & 12.00         & 12.00       & \B{1.00} & 11.67    & 12.00    & 9.40                      & 11.08       & 11.92        \\
        Cosine        & 12.00           & \ul{3.00}     & \B{2.00}    & 10.00    & 7.67     & 7.25     & 6.99                      & 6.38        & 4.98         \\
        Hyb MSE       & 6.00            & 11.00         & \ul{3.00}   & \ul{2.00}& 8.83     & 8.25     & 6.51                      & 7.42        & 7.77         \\
        Hyb SmL1      & 8.00            & 5.00          & 4.50        & 9.00     & 5.17     & 6.00     & 6.28                      & 5.73        & 5.17         \\
                    \hline   
                    & \multicolumn{9}{c}{\B{Standardization}}                                                                                                       \\
                    \hline
        Global Stdze  & \ul{2.75}       & 5.00          & 4.50        & 5.00     & \ul{4.33}& 4.50     & \ul{4.35}                 & \ul{4.22}   & \ul{4.58}    \\
        Standardize   & \ul{2.75}       & 7.00          & 7.50        & 3.00     & \B{3.83} & 4.75     & 4.81                      & 5.17        & 5.77         \\
        \B{PHI-S}          & \B{2.00}        & 5.00          & \B{2.00}    & 4.00     & \ul{4.33}& \B{3.00} & \B{3.39}                  & \B{3.27}    & \B{3.58}     \\
                    \hline   
                    & \multicolumn{9}{c}{\B{Whitening}}                                                                                                             \\
                    \hline
        PCA-W         & 7.75            & 9.50          & 7.50        & 8.00     & 7.33     & \ul{3.75}& 7.31                      & 7.17        & 7.02         \\
        ZCA           & 6.75            & 9.50          & 8.00        & 11.00    & 5.67     & 4.50     & 7.57                      & 6.88        & 6.92         \\
        HCA           & 8.00            & 8.00          & 6.00        & 12.00    & 7.33     & 5.00     & 7.72                      & 6.87        & 6.58         \\
                     \hline                                                                                                                                               
                     & \multicolumn{9}{c}{\B{AdaLoss}}                                                                                                              \\
                     \hline                                                                                                                                             
        MSE          & 7.75             & \B{1.50}      & 11.00       & 7.00     & 5.83     & 9.25     & 7.06                      & 7.07        & 6.90         \\           
        PHI-S         & 6.25             & \B{1.50}      & 10.00       & 6.00     & 6.00     & 9.75     & 6.58                      & 6.70        & 6.81         
    \end{tabular}
    }
    \caption{Average benchmark ranks across the suite including AdaLoss. For LLaVA, we first average the two GQA and TextVQA tasks separately, and then combine those with POPE and VQAv2 to compute the average. This is to prevent overly biasing towards the tasks that have multiple measurements. We observe that the standardization techniques perform the best, with PHI-S being the strongest normalization method studied. AdaLoss was able to improve over baseline, but is not competitive with the standardization methods. The raw benchmark scores are provided in appendix \ref{sec:vitb_raw_scores}.}
    \label{tab:vitb_method_ranks_adaloss}
\end{table*}

\subsubsection{ViT-B/16}\label{sec:vitb_raw_scores}

In table \ref{tab:vitb_raw_classification_semseg} we show the raw benchmark scores for classification, segmentation, and Probe 3D \cite{elbanani2024probing}. When viewing the raw scores, it's less clear what the ideal method is, if any, aside from it being fairly obvious that the MSE baseline is the worst. We also show the metrics for LLaVA 1.5 integration in \ref{tab:vitb_raw_llava}. It's easiest to see the best performing method by looking at the average ranks across the task suite in table \ref{tab:vitb_method_ranks}, where being consistently strong is more evident. The ``Ada -'' prefix means that we used AdaLoss.

\begin{table}[!ht]
    \centering
    \resizebox{\linewidth}{!}{
    \begin{tabular}{r|cc|ccc|cccc}
                                & \multicolumn{2}{c|}{\B{Classification}} & \multicolumn{3}{c|}{\B{Segmentation}} & \multicolumn{4}{c}{\B{Probe 3D}}       \\
                                &               &         &            &            & \B{SAM}     &            & \B{Surface}  & \B{Multi-}  & \B{SPair}  \\
        \B{Method} $\uparrow$   & \B{Zero Shot} & \B{kNN} & \B{ADE20k} & \B{VOC}    & \B{COCO}    & \B{Depth}  & \B{Normals}  & \B{View}    & \B{71k}    \\
        \hline                                                                                       
        MSE           & 56.17          & 71.54            & 42.40      & 78.10      & \B{71.90}   & 77.69      & 55.06        & 47.71       & 33.56      \\
        Cosine        & 71.44          & 79.74            & 48.01      & \B{83.39}  & 69.42       & 81.77      & 56.46        & 53.53       & 39.59      \\
        Hyb MSE       & 69.34          & 78.72            & 48.00      & \ul{83.29} & \ul{70.54}  & 80.88      & 56.30        & 52.57       & 43.44      \\
        Hyb SmL1      & 71.19          & 79.49            & \ul{48.23} & 82.82      & 69.53       & \B{82.14}  & 56.43        & 53.69       & 40.45      \\
        \hline                                                                                       
        Global Stdze  & 70.91          & 79.51            & 47.89      & 83.07      & 69.75       & \ul{82.02} & 57.02        & 54.13       & 42.53      \\
        Standardize   & 70.51          & 79.35            & 47.87      & 82.79      & \ul{70.22}  & 80.44      & 56.48        & \B{54.65}   & \B{45.27}  \\
        PHI-S          & 70.73          & 79.53            & \B{48.63}  & 83.09      & 69.89       & 81.89      & 56.79        & \ul{54.49}  & 43.92      \\
        \hline                                                                                                                                           
        PCA-W         & 70.23          & 79.30            & 47.58      & 82.96      & 69.55       & 81.88      & 56.71        & 54.42       & \ul{44.24} \\
        ZCA           & 70.38          & 79.28            & 47.83      & 82.80      & 69.37       & 81.43      & \B{57.23}    & \ul{54.49}  & 43.35      \\
        HCA          & 70.47          & 79.33            & 47.84      & 82.99      & 69.19       & 81.61      & \ul{57.07}   & 54.35       & 43.14      \\
        \hline
        Ada - MSE     & \B{72.89}      & \ul{79.85}       & 47.24      & 82.53      & 69.60       & 81.57      & 56.33        & 51.86       & 36.85      \\
        Ada - PHI-S    & \ul{72.73}     & \B{80.03}        & 47.41      & 82.72      & 69.74       & 81.72      & 55.49        & 51.28       & 36.46
    \end{tabular}
    }
    \caption{\B{ViT-B/16} - Classification accuracy using both Zero Shot (DFN CLIP text encoder) and kNN. ADE20k and VOC are semantic segmentation linear probe results using 512px resolution (see \cite{ranzinger2023amradio} for details), and SAM COCO instance segmentation, also defined in AM-RADIO. We also show the Probe 3D \cite{elbanani2024probing} metrics as also reported in AM-RADIO.}
    \label{tab:vitb_raw_classification_semseg}
\end{table}

\begin{table*}[!ht]
    \centering
    \resizebox{0.75\linewidth}{!}{
    \begin{tabular}{r|cc|cc|c|c}
                     & \multicolumn{2}{c|}{\B{GQA}} & \multicolumn{2}{c|}{\B{TextVQA}} &           &            \\
        \B{Method} $\uparrow$  & \B{Val}  & \B{TestDev}     & \B{Tokens} & \B{No Tokens}     & \B{POPE} & \B{VQAv2} \\
        \hline
        MSE          & 67.35     & 59.51            & 47.31       & 15.06              & 85.16     & 72.21      \\
        Cosine       & 70.02     & 61.82            & 50.24       & 24.13              & 84.78     & 76.14      \\
        Hyb MSE      & 69.86     & 61.96            & 50.15       & 23.53              & 85.19     & 75.94      \\
        Hyb SmL1     & 70.03     & 62.35            & 50.19       & 23.90              & 85.74     & 76.17      \\
        \hline
        Global Stdze & \ul{70.10}& 62.28            & 50.31       & 22.55              & \ul{85.88}& \ul{76.21} \\
        Standardize  & 70.04     & 62.16            & 50.28       & 24.20              & \B{85.94}& 76.20       \\
        PHI-S         & \B{70.20} & \B{62.55}        & 50.25       & 23.28              & 85.52     & \B{76.30}  \\
        \hline                                                                                                  
        PCA-W        & 69.85     & 62.01            & 50.48       & 24.14              & 85.43     & 75.93      \\
        ZCA          & 69.98     & \ul{62.37}       & 50.11       & 24.63              & 85.80     & 76.02      \\
        HCA          & 69.95     & 61.79            & 49.92       & 24.79              & 85.61     & 76.07      \\
        \hline
        Ada - MSE    & 69.75     & 62.31            & \ul{50.82}  & \B{26.63}          & 85.06     & 76.09      \\
        Ada - PHI-S   & 69.76     & 61.90            & \B{50.90}   & \ul{25.81}         & 85.54     & 76.03
    \end{tabular}
    }
    \caption{\B{ViT-B/16} - LLaVA 1.5 (Vicuna 7B) results. We use the same suite in AM-RADIO, however we report both ``Val'' and ``TestDev'' for GQA, and also report the TextVQA score when OCR tokens are not provided as part of the context.}
    \label{tab:vitb_raw_llava}
\end{table*}

\subsubsection{ViT-L/16}\label{sec:vitl_raw_scores}

In table \ref{tab:raw_mse_vitl} we show the MSE for our ViT-L/16 trained student model. Similar to the ViT-B/16 metrics, PHI-S does the best job of simultaneously minimizing all of the teacher errors. We also provide the raw benchmark scores in tables \ref{tab:vitl_raw_classification_semseg} and \ref{tab:vitl_raw_llava}.

\begin{table}[!ht]
    \centering
    \resizebox{0.7\linewidth}{!}{
    \begin{tabular}{r|cccc}
        \B{Method} $\downarrow$   & \B{DFN CLIP} ($\cdot 1^{-4}$)    & \B{SigLIP} & \B{DINOv2} & \B{SAM}    \\
        \hline
        Baseline - MSE            & 5.0200                           & 1.9030     & 0.9591     & \B{5.9970}\\
        Global Stdze              & 4.6640                           & 1.8620     & \B{0.6924} & 7.9080 \\
        Standardize               & \ul{4.6520}                      & \ul{1.8560}& 0.7036     &\ul{7.7030}\\
        \B{PHI-S}                 & \B{4.6310}                       & \B{1.8460} & \ul{0.6961}& 7.7190
    \end{tabular}
    }
    \caption{\B{ViT-L/16} - Mean Squared Error for matching the teachers different algorithms. Lower values are better.}
    \label{tab:raw_mse_vitl}
\end{table}

\begin{table}[!ht]
    \centering
    \resizebox{\linewidth}{!}{
    \begin{tabular}{r|cc|ccc|cccc}
                                & \multicolumn{2}{c|}{\B{Classification}} & \multicolumn{3}{c|}{\B{Segmentation}} & \multicolumn{4}{c}{\B{Probe 3D}}       \\
                                &               &            &            &            & \B{SAM}     &            & \B{Surface}  & \B{Multi-}  & \B{SPair}  \\
        \B{Method} $\uparrow$   & \B{Zero Shot} & \B{kNN}    & \B{ADE20k} & \B{VOC}    & \B{COCO}    & \B{Depth}  & \B{Normals}  & \B{View}    & \B{71k}    \\
        \hline                                                                                       
        Baseline - MSE          & 71.32         & 78.80      & 47.01      & 82.62      & \B{72.91}   & 80.21      & 57.50        & 48.44       & 35.67      \\
        Global Stdze            & 78.59         & \ul{83.15} & 50.94      & \ul{85.58} & 71.23       & \ul{84.51} & 60.27        & 57.86       & \ul{52.24} \\
        Standardize             & \ul{78.67}    & 83.05      & \B{51.27}  & 84.79      & \ul{71.69}  & 84.04      & 60.27        & \B{58.34}   & \B{52.42}  \\
        PHI-S                    & \B{78.68}     & \B{83.16}  & \ul{51.23} & \B{85.73}  & 71.12       & \B{84.77}  & \B{60.61}    & \ul{58.22}  & 51.74
    \end{tabular}
    }
    \caption{\B{ViT-L/16} - Classification accuracy using both Zero Shot (DFN CLIP text encoder) and kNN. ADE20k and VOC are semantic segmentation linear probe results using 512px resolution (see \cite{ranzinger2023amradio} for details), and SAM COCO instance segmentation, also defined in AM-RADIO. We also show the Probe 3D \cite{elbanani2024probing} metrics as also reported in AM-RADIO.}
    \label{tab:vitl_raw_classification_semseg}
\end{table}

\begin{table*}[!ht]
    \centering
    \resizebox{0.75\linewidth}{!}{
    \begin{tabular}{r|cc|cc|c|c}
                     & \multicolumn{2}{c|}{\B{GQA}} & \multicolumn{2}{c|}{\B{TextVQA}} &           &            \\
        \B{Method} $\uparrow$  & \B{Val}   & \B{TestDev}     & \B{Tokens} & \B{No Tokens}     & \B{POPE}  & \B{VQAv2} \\
        \hline
        Baseline - MSE         & 69.70     & 62.11           & 48.88      & 21.21             & 85.72     & 75.44     \\
        Global Stdze           & \B{71.65} & \ul{63.08}      & \B{53.15}  & 31.43             & 86.03     & \B{78.37} \\
        Standardize            & 71.44     & \B{63.11}       & \ul{52.97} & \ul{33.56}        & \ul{86.21}& 78.19     \\
        PHI-S                   & \ul{71.46}& 63.07           & 52.88      & \B{33.67}         & \B{86.29} & \ul{78.31}
    \end{tabular}
    }
    \caption{\B{ViT-L/16} - LLaVA 1.5 (Vicuna 7B) results. We use the same suite in AM-RADIO, however we report both ``Val'' and ``TestDev'' for GQA, and also report the TextVQA score when OCR tokens are not provided as part of the context.}
    \label{tab:vitl_raw_llava}
\end{table*}

\subsection{Comparison with recent Agglomerative Models}

Along with AM-RADIO at CVPR, Theia~\cite{shang2024theia} has been published to CoRL, and there are recent preprints for UNIC~\cite{sariyildiz2024unicuniversalclassificationmodels}, and UNIT~\cite{zhu2024unit}. We report benchmarks that are common amongst the papers in table \ref{tab:agglomerative_comparison}, but note that only AM-RADIO and Theia are published, and thus the other works are potentially subject to change as they work through the peer review process. For each model, we report the numbers from the original papers without attempting replication. We do run linear probing for Theia on the ADE20k task using our harness as it allows for the only task that all papers report. We confirmed the mIoU numbers with the authors before reporting them here. We also note that the settings, such as training dataset, resolution, set of teachers, and desired outcomes, are different between the models, which means there are numerous confounding factors preventing the comparison from being ``fair''.

\begin{table}[!ht]
    \centering
    \resizebox{0.7\linewidth}{!}{
    \begin{tabular}{rc|ccc|c}
        \multirow{2}{*}{\B{Method}}            & \multirow{2}{*}{\B{Model}} & \multicolumn{3}{c|}{\B{ImageNet-1K Classification}} & \B{Segmentation} \\
                                               &                            & \B{Zero Shot} & \B{kNN}   & \B{Probe}   & \B{ADE20k}       \\
        \hline                                                                                       
        AM-RADIO                               & ViT-H/16                   & \B{82.93}     & \B{86.06} & -           & 51.34            \\
        Theia                                  & ViT-B/16                   & -             & -         & 75.2        & 35.61*           \\
        UNIC                                   & ViT-B/16                   & -             & -         & \B{83.2}    & 37.3             \\
        UNIT                                   & ViT-H/14                   & 78.76         & 84.18     & -           & 50.19            \\
        \hline
        \multirow{2}{*}{PHI-S-RADIO}            & ViT-B/16                   & 73.16         & 81.74     & -           & 48.94            \\
                                               & ViT-L/16                   & 80.45         & 84.57     & -           & \B{51.47}
    \end{tabular}
    }
    \caption{
    Comparison between shared metrics of different agglomerative model approaches.
    \newline
    *Our results
    }
    \label{tab:agglomerative_comparison}
\end{table}

\end{document}